\documentclass[10pt]{IEEEtran}
\usepackage{amsfonts}
\usepackage{amsmath}
\usepackage{amssymb}
\usepackage{bm}
\usepackage{cite}
\usepackage{graphicx}
\usepackage{epstopdf}
\usepackage{epsfig}
\usepackage{url}
\usepackage{setspace}
\usepackage{subfigure}
\usepackage{color}
\usepackage{algorithmicx}
\usepackage{algpseudocode}
\usepackage{amsmath}
\usepackage{algorithm,times}
\usepackage{threeparttable}
\usepackage{lineno}
\usepackage{multirow}
\usepackage{verbatim}
\usepackage{booktabs}
\usepackage{stfloats}
\usepackage{makecell}
\graphicspath{{./figures/}}

\begin{document}

\title{ 
	Improving the Anomaly Detection in GPR Images by Fine-Tuning CNNs with Synthetic Data
	\thanks{%
Xiren Zhou, Shikang Liu, Ao Chen, Yizhan Fan, and Huanhuan Chen are with School of Computer Science
and Technology, University of Science and Technology of China, Hefei,
230027, China; e-mail: zhou0612@mail.ustc.edu.cn,  qq1321401109@mail.ustc.edu.cn, chenao57@mail.ustc.edu.cn, fyz666@mail.ustc.edu.cn, hchen@ustc.edu.cn. \emph{Corresponding Author: Huanhuan Chen}}
\author{Xiren Zhou, Shikang Liu, Ao Chen, Yizhan Fan,  Huanhuan~Chen,~\IEEEmembership{Senior Member,~IEEE}}
}
\maketitle

\begin{abstract}
Ground Penetrating Radar (GPR) has been widely used to estimate the healthy operation of some urban roads and underground facilities. When identifying subsurface anomalies by GPR in an area, the obtained data could be unbalanced, and the numbers and types of possible underground anomalies could not be acknowledged in advance. In this paper, a novel method is proposed to improve the subsurface anomaly detection from GPR B-scan images. A normal (i.e. without subsurface objects) GPR image section is firstly collected in the detected area. 
Concerning that the GPR image is essentially the representation of electromagnetic (EM) wave and propagation time, and to preserve both the subsurface background and objects' details, the normal GPR image is segmented and then fused with simulated GPR images that contain different kinds of objects to generate the synthetic data for the detection area based on the wavelet decompositions.
Pre-trained CNNs could then be fine-tuned with the synthetic data, and utilized to extract features of segmented GPR images subsequently obtained in the detection area. The extracted features could be classified by the one-class learning algorithm in the feature space without pre-set anomaly types or numbers.
The conducted experiments demonstrate that fine-tuning the pre-trained CNN with the proposed synthetic data could effectively improve the feature extraction of the network for the objects in the detection area. Besides, the proposed method requires only a section of normal data that could be easily obtained in the detection area, and could also meet the timeliness requirements in practical applications.
\end{abstract}
\begin{IEEEkeywords}
	Ground Penetrating Radar, B-scan Image, Anomaly Detection, Buried Asset
	Detection, Data Processing.	
\end{IEEEkeywords}

\section{Introduction}

\begin{figure}[htbp]
	\centering
	\subfigure[]{ \centering
		\includegraphics[height=0.9in]{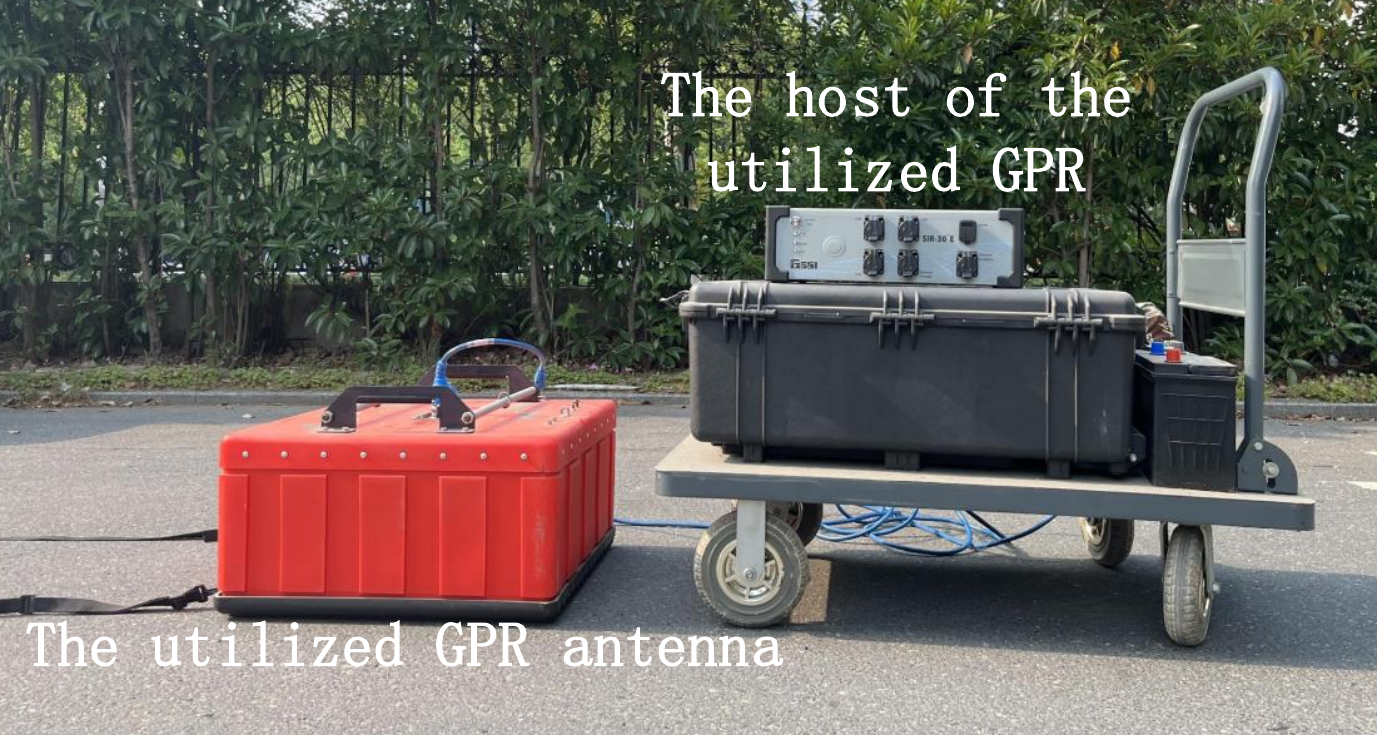}}
	\subfigure[]{ \centering
		\includegraphics[height=0.9in]{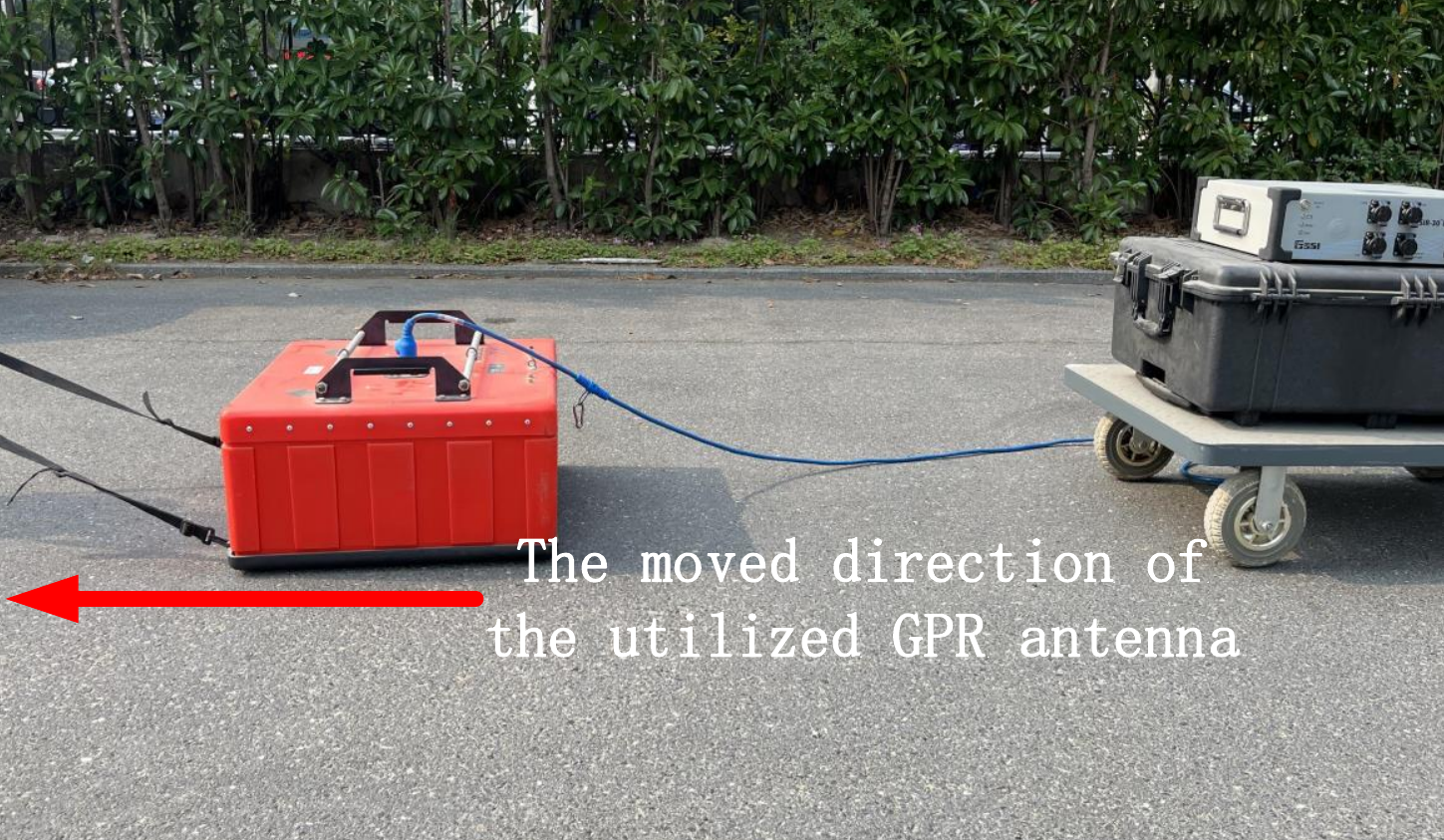}}
	\subfigure[]{ \centering
		\label{abscan}
		\includegraphics[height=1.2in]{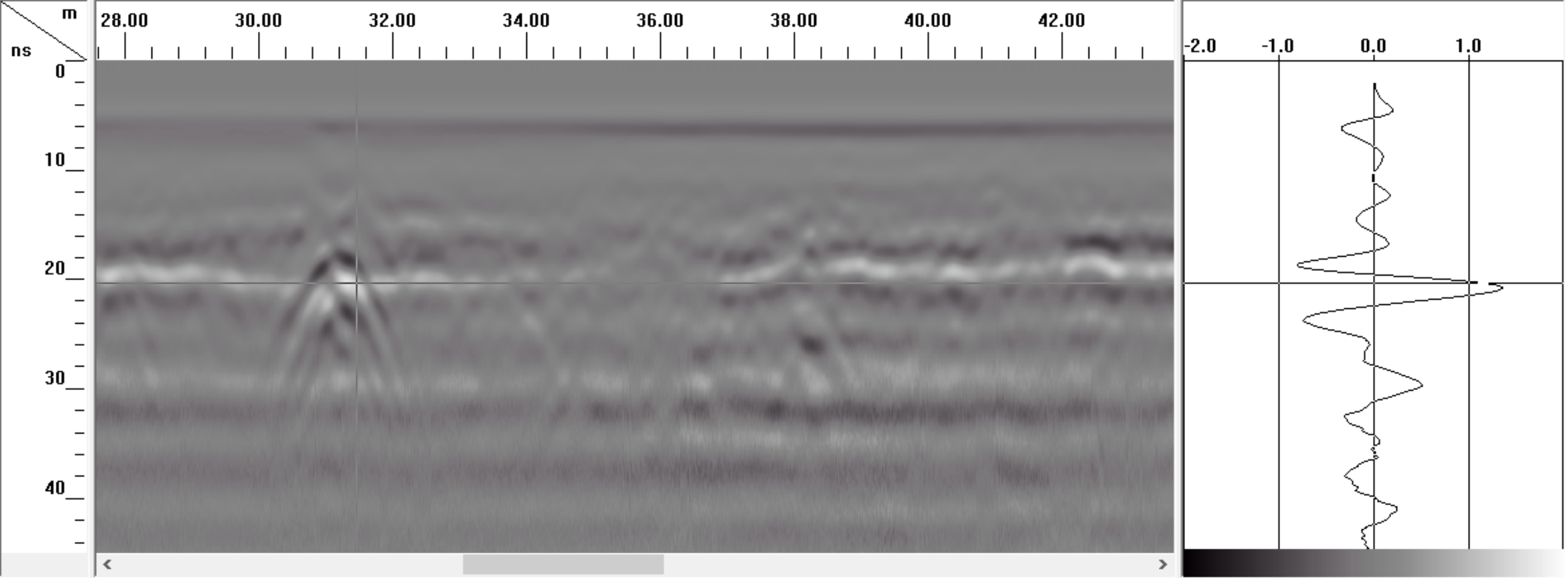}}
	\caption{ (a) and (b) present a real detection along an urban road using GPR. (a) shows the host and antenna of the utilized GSSI GPR. (b) shows the actual detection scene of a  road. The GPR antenna is moved along the road (the red arrow) to obtain the GPR B-scan image, and the collected image is transformed to the host. (c) illustrates the generation of a GPR B-can image. The abscissa in the left B-scan image is the detection position, and the ordinate is the time from emission to reception of the electromagnetic wave. The right side is a received wave in the left B-scan image. As shown in the bottom of the right side,  different gray values correspond to different magnitudes of the EM wave. The gray value of each pixel represents the corresponding amplitude at this position and time.
	}
	\label{gprtest}
\end{figure}

Ground Penetrating Radar (GPR) has become increasingly important as a nondestructive tool to estimate the healthy operation of some urban roads and underground facilities, which makes the use of the transmission and reflection of the electromagnetic (EM) waves to detect dielectric properties changes in host materials \cite{daniels2004ground}. By arranging received EM waves horizontally in a temporal or spatial relationship, and representing the wave intensities with corresponding grayscale values, a GPR B-scan image could be obtained \cite{neal2004ground} as shown in Fig. \ref{gprtest}.
The subsurface situation or existing objects could be estimated by interpreting different shapes or characteristics on the obtained GPR B-scan image\cite{chen2010buried}. But due to the refraction and reflection of EM waves, the various shapes on the GPR image\footnote{The GPR image in this paper refers to the GPR B-scan image as shown in Fig. \ref{abscan}.} are not actually the same as the actual objects. Besides, the system noise, the heterogeneity of the medium, and mutual wave interactions also make it challenging to automatically cope with the GPR images obtained in the detection area\cite{hao2020air}.

\begin{figure*}[htpb]
	\centering
	\includegraphics[width=0.995\textwidth]{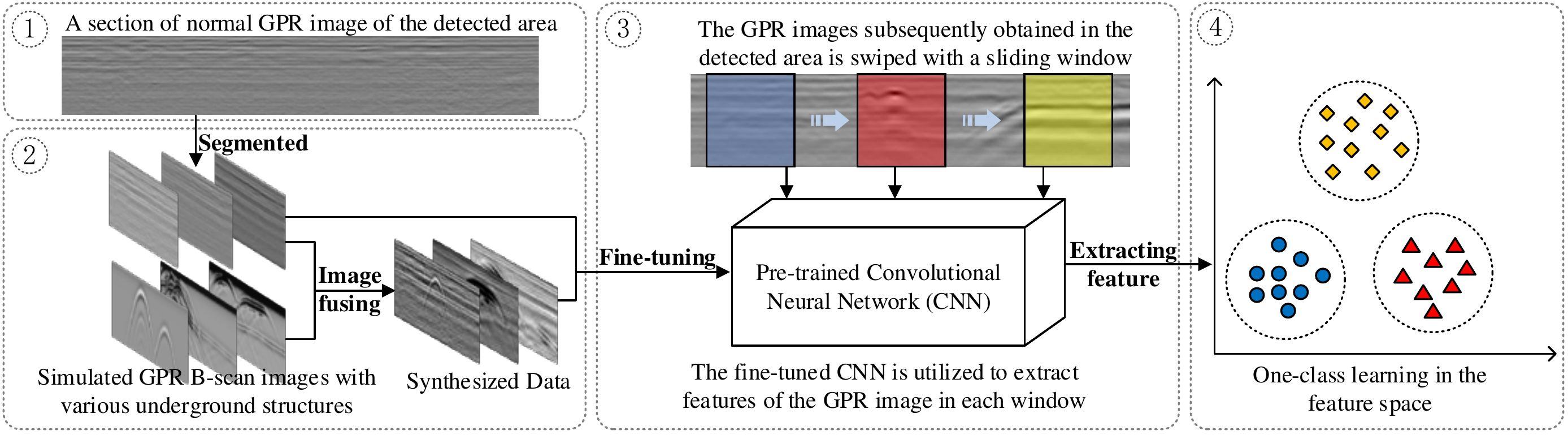}
	\caption{This figure shows the procedure of the proposed method, which could be divided into four steps.
	1) A section of a normal GPR image without any subsurface objects is obtained in the detection area.  
	2) The normal GPR image section is segmented, and fused with simulated GPR images containing various kinds of subsurface objects to generate synthesized data for the detection area. 
	3) The pre-trained CNN is fine-tuned with the generated synthesized data. The subsequently obtained GPR image is swiped with a sliding window, and the fine-tuned CNN is utilized to extract features of the GPR image in each window. 
	4) The extracted features are finally classified by the one-class learning algorithm in the feature space, and the corresponding underground anomalies are detected.}
	\label{Procedure}
\end{figure*}

The interpretation of GPR B-scan images could be roughly grouped into two categories: extracting and fitting hyperbolic characteristics on B-scan images, and identifying non-hyperbolic shapes.
The hyperbolic characteristics are generated by underground objects with circular cross sections (e.g. tree roots, pipes, etc.).
In addition to hyperbolic characteristics, non-hyperbolic shapes could be more common targets in GPR B-scan images when imaging the subsurface. These shapes could be formed by different kinds of subsurface media or objects, including subsurface cavities, moisture damage, loose media, etc. Different subsurface media or objects would generate shapes with different characteristics on the GPR image. Even for the same type of underground objects, the generated shapes on the GPR image vary in different underground environments or by different GPR\cite{chen2011buried}.

To identify non-hyperbolic shapes in GPR images, signal or image processing methods \cite{liu20203d,rasol2020gpr}, and Convolutional Neural Network (CNN) \cite{tong2018innovative,tong2020pavement,du2021pavement} are utilized. But there are still some practical issues that need to be considered when identifying underground anomalies from GPR images obtained in an unknown area: 1) The obtained data could be unbalanced, since the amount of GPR image generated by subsurface anomalies could be much smaller than the normal data, and there could be even no abnormality in the detection area. 2) Only some normal GPR images without any underground objects could be obtained at the beginning of the detection, and it could be a common situation that the numbers and types of the possible underground anomalies are rarely acknowledged prior to fully analyzing the obtained GPR data. 3) The underground environments vary in areas, thus there is no guarantee that training a model with data from an area or datasets will improve its performance in other areas. 4) In real-world applications, there could be timeliness requirements for anomaly detection, and the locations of anomalies need to be marked on-site for further repairs.

To address the above issues, a novel method based on CNN and one-class learning is proposed in this paper to improve the subsurface anomaly detection from the obtained GPR images. The procedure of the proposed method is presented in Fig. \ref{Procedure}, which could be roughly divided into four steps: 
1) A section of a normal GPR image without any subsurface objects is obtained in the detection area;  
2) The normal GPR image section is segmented, and fused with simulated GPR images containing various kinds of subsurface objects to generate synthesized data for the detection area; 
3) The pre-trained CNN is fine-tuned with the generated data, and used to extract features of the subsequently obtained GPR images; 
4) The one-class learning algorithm is utilized to identify anomalies from the extracted features.

Existing studies \cite{pan2009survey,weiss2016survey,zhu2021adaptive,chen2009probabilistic} on transfer learning have shown that proper transfer of networks could effectively improve the learning results in the case of insufficient training data, and any pre-trained CNN could be adopted as a feature extractor for further feature learning \cite{oza2018one,pham2018buried,chen2013learning}. 
The features extracted from different inputs constitute a feature space, and the distance between two features in the feature space reflects the difference between the two corresponding inputs. 
To improve the CNN's ability to identify different types of inputs in a specific task, that is, to increase the inter-class distance in the feature space, the pre-trained CNN could be further trained or fine-tuned with the data that contains both the background and target details of this task \cite{radenovic2018fine,li2022enhanced}. 
Pre-trained CNNs with adequate fine-tuning could perform at least as well as full training, and the dependence on the amount of data could be significantly reduced \cite{tajbakhsh2016convolutional}.
However, as aforementioned, when detecting an area, only a section of a normal GPR image could be obtained, which is not enough to provide the type and amount of data for fine-tuning. Unlike visual images, GPR images have their own physical meaning. The conducted experiments one real-world dataset show limited effects of the feature extraction of GPR images using CNNs only trained on visual images. 

Nonetheless, by utilizing appropriate image fusion techniques, remote sensing images that are very close to real ones could be synthesized, and these generated data could be used for change detection and object recognition\cite{zhang2022adaptive,quan2022deep,zhang2022sparse}. The GPR simulation software such as GprMax \cite{giannopoulos2005modelling} could generate GPR images of various underground objects in pre-set environments.
Therefore, in this paper, a section of a normal GPR image is firstly obtained in the detection area, and then fused with some simulated GPR images containing various kinds of underground objects to generate synthetic data specifically designed for the subsurface environment of this area. 
Concerning that the GPR image is essentially the representation of EM wave intensity and propagation time as Fig. \ref{abscan}, and to preserve both the subsurface background and target details, the wavelet decompositions of the normal and simulated images are merged to generate the synthetic images that contain both the basic underground conditions of the detection area and various characteristics generated by underground objects. The conducted experiments demonstrate that fine-tuning pre-trained CNNs with the generated synthetic data could improve the feature extraction of the network for further learning, i.e. features extracted from different kinds of objects have larger distances in the feature space.

Due to the unknown numbers and types of subsurface anomalies that may arise, there should be no preset anomaly types in real-world applications. In the proposed method, a sliding window is constructed and swiped across the obtained GPR image. The normal GPR B-scan images in the window without any underground objects are mapped to the feature space via the fine-tuned CNN, and an initial one-class classifier is trained. For the subsequent features extracted from the image in the sliding window, the trained classifier is continuously used for classification. The abnormal data is further trained by incremental one-class learning, where more incremental classifiers could be obtained to classify the features into classes. Thus the corresponding underground object that generates the GPR image could be detected.

The main contributions of the proposed method could be summarized as follows.
\begin{enumerate}
	\item A section of a normal GPR image is obtained in the detection area, segmented, and fused with simulated GPR images to generate synthetic data that contains both the basic underground conditions of the detection area and various characteristics generated by underground objects. The conducted experiments demonstrate that fine-tuning CNNs with the synthetic data could increase the distance between features extracted from GPR images formed by different objects in the detection area.
	\item Only a section of a normal GPR image of the detection area (without subsurface anomalies) is required to perform the proposed method, instead of an amount of normal and abnormal data that is difficult to collect, process, and label in real-world applications.
	\item By performing one-class learning, there is not need to know the number and types of subsurface anomalies that may exist in the detection area in advance. The proposed method could incrementally classify various types of underground objects through the extracted features.
\end{enumerate}

The rest of this paper is organized as follows. Some related work about interpreting GPR images is presented in Section II. The feature extraction is presented in Section III, including generating synthetic images, and fine-tuning a pre-trained CNN. The one-class learning is introduced in Section IV. Experiments are conducted and analyzed in Section V. Finally, conclusions are drawn in Section VI.

\section{Related Work}
Interpreting GPR B-scan images could be roughly grouped into two categories: extracting and fitting hyperbolic characteristics on B-scan images,  and identifying non-hyperbolic shapes.
The prevailing methodologies in hyperbolic recognition from GPR images include the Hough transform (HT)\cite{porrill1990fitting,borgioli2008detection,windsor2014data}, Machine Learning (ML)\cite{maas2013using,jiang2019cable,jiang2019} and some methods that combine multiple approaches\cite{chen2010probabilistic,chen2010robust,dou2016real,zhou2019probabilistic,li2019toward}. In our previous work\cite{zhou2018automatic}, a GPR B-scan image interpreting model has been proposed, which could estimate the radius and depth of the buried pipelines by extracting and fitting  hyperbolic point clusters from GPR B-scan images.

Besides hyperbolic characteristics, some existing studies identify underground objects with non-hyperbolic shapes from GPR data by signal processing or image recognition methods. The frequency-domain-focusing (FDF) technology of synthetic aperture radar (SAR) is utilized to aggregate scattered GPR signals for acquiring testing images, where a low-pass filter is designed to denoise primordial signals, and the profiles of detecting objects are extracted via the edge detection technique based on the background information \cite{liu20203d}. Subsequently, a formula is conducted to relate the hidden crack width with the relative measured amplitude \cite{rasol2020gpr}. The use of this kind of method generally requires knowledge of the basic conditions of the underground medium in advance.

To locate and identify objects in GPR images, the Convolutional Neural Network (CNN) is utilized in recent decades. 
Regarding the EM signals as an input value, CNN structures are conducted to automatically localize several kinds of targets in GPR data \cite{tong2020pavement}. The You-Only-Look-Once (YOLO) \cite{ukhwah2019asphalt} is also utilized to detect potholes and crackings beneath the roads. 
Zhang \emph{et al.} propose a mixed deep CNN model combined with the Resnet-50 base network and YOLO framework to detect the moisture damage in GPR data. Subsequently, Liu \emph{et al.} propose a method combining the YOLO series with GPR images to recognize the internal defects in asphalt pavement \cite{liu2021application}.  When detecting the underground objects in a certain area, it is difficult to ensure that the existing training data of CNN obtained from other areas or datasets is consistent with the underground situation in this area or road. And only some GPR images without any target objects could be obtained at the beginning of the detection, resulting in insufficient training data of CNN. The Generative Adversarial Network (GAN) \cite{ren2020unsupervised} could be utilized to generate remote-sensing data, but the main issue with this approach in our scenario is that training the generative network for the detection area could be time-consuming for on-site applications.

\section{Feature Extraction by Fine-tuned CNNs}

In this section, the generation of synthetic data of the detection area is firstly introduced. After that, the pre-trained CNN is fine-tuned with the synthetic and normal data to enhance the feature extraction capabilities for the objects in the detection area.

\subsection{Generating the Synthetic Data for the detection area}

\subsubsection{The Data Sources}

As aforementioned, the synthetic data is fused from two sources: 1) The normal GPR image obtained in the detection area without any underground objects; 2) The GPR images simulated by GprMax with various kinds of buried objects. 

When detecting an area (e.g. a pavement road), a GPR image section without any buried objects could be easily obtained. This image section could be used to describe the basic subsurface environment of the detection area, and provide data for CNN to extract the features of the GPR image in the area without underground objects. In the conducted experiments of this paper, the GPR image section with a length greater than 3000 pixels is collected, and more than 300 GPR image segment with the horizontal length of 300 pixels is then randomly selected in this GPR image section \footnote{Duplications could exist in the selected images.}.

\begin{figure*}[h]
	\centering
	\subfigure[]{ \centering
		\label{n1}
		\includegraphics[width=0.154\textwidth]{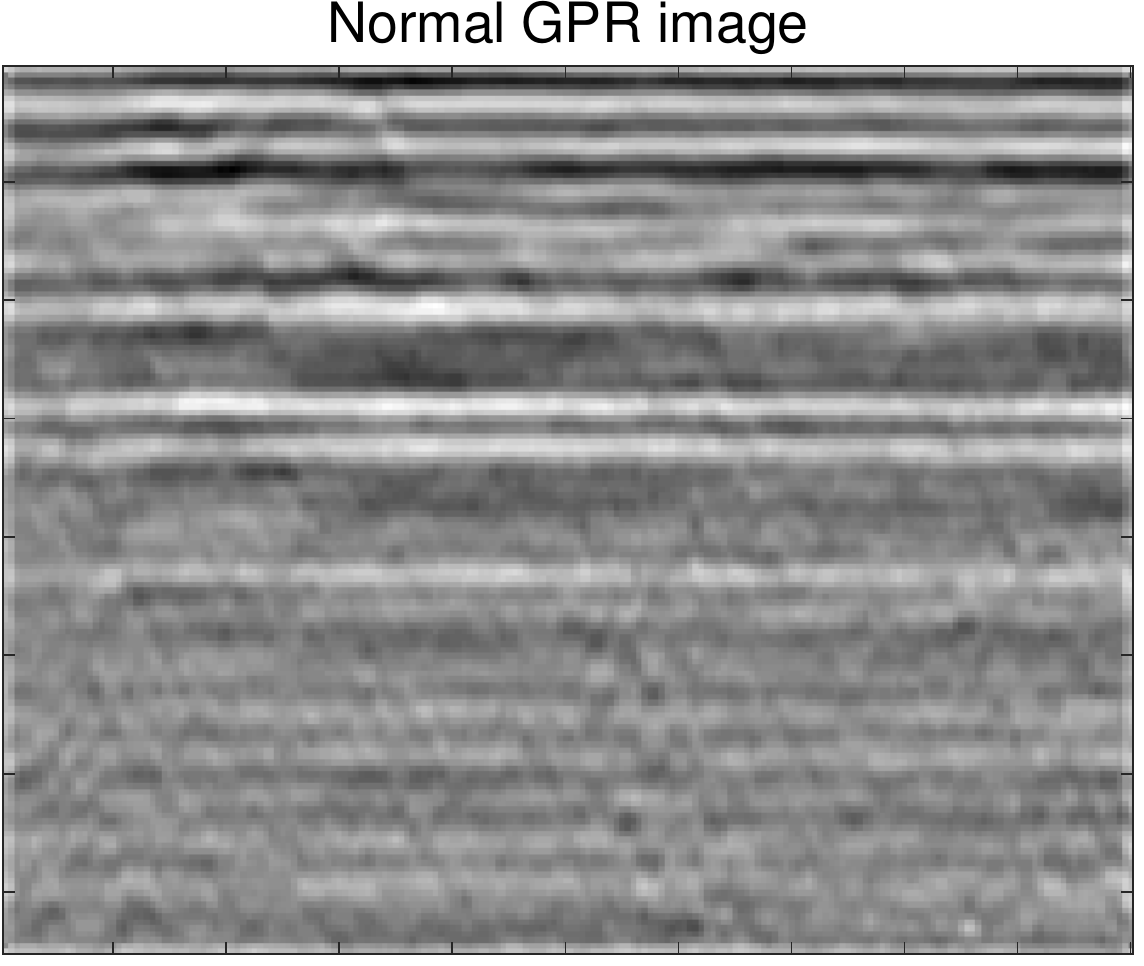}}
	\subfigure[]{ \centering
		\label{n5}
		\includegraphics[width=0.154\textwidth]{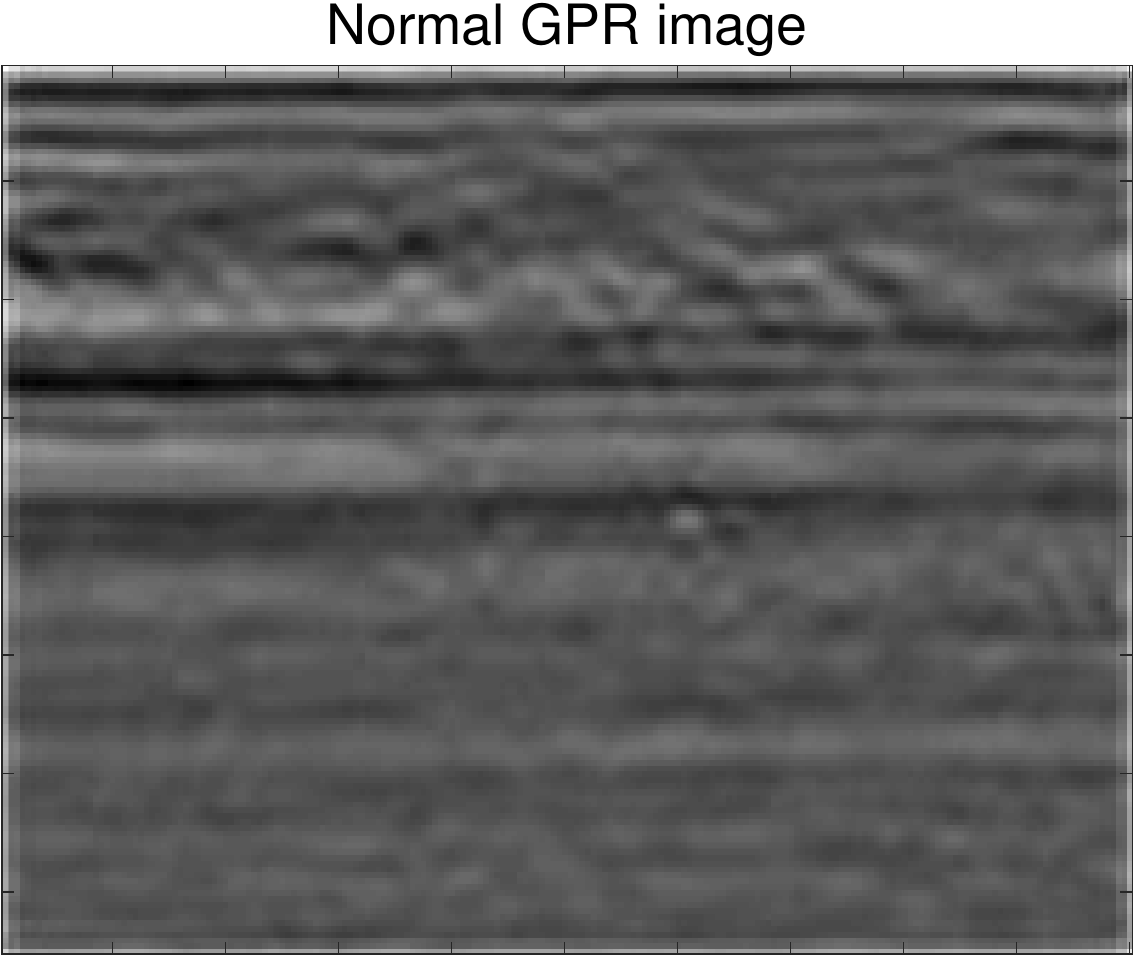}}
	\subfigure[]{ \centering
		\label{n2}
		\includegraphics[width=0.154\textwidth]{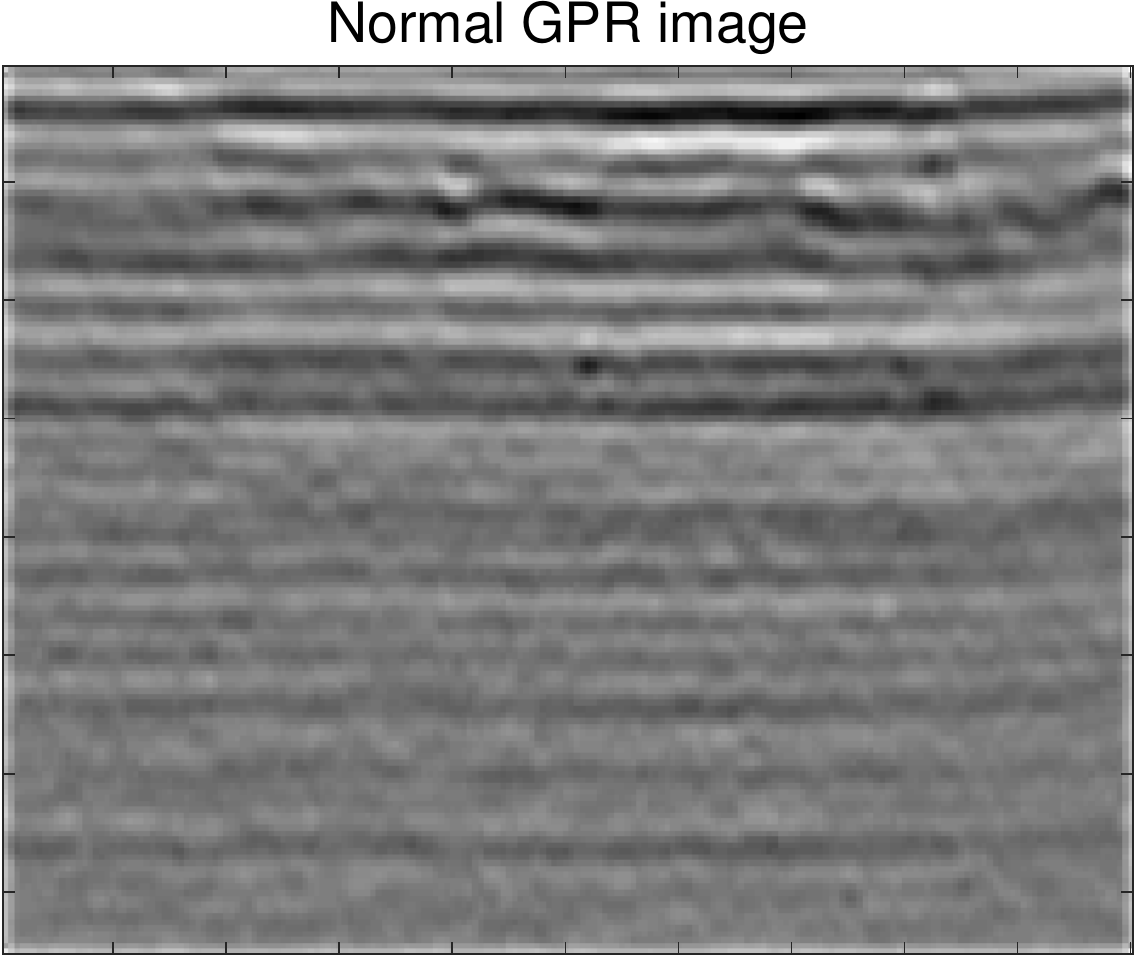}}
	\subfigure[]{ \centering
		\label{n3}
		\includegraphics[width=0.154\textwidth]{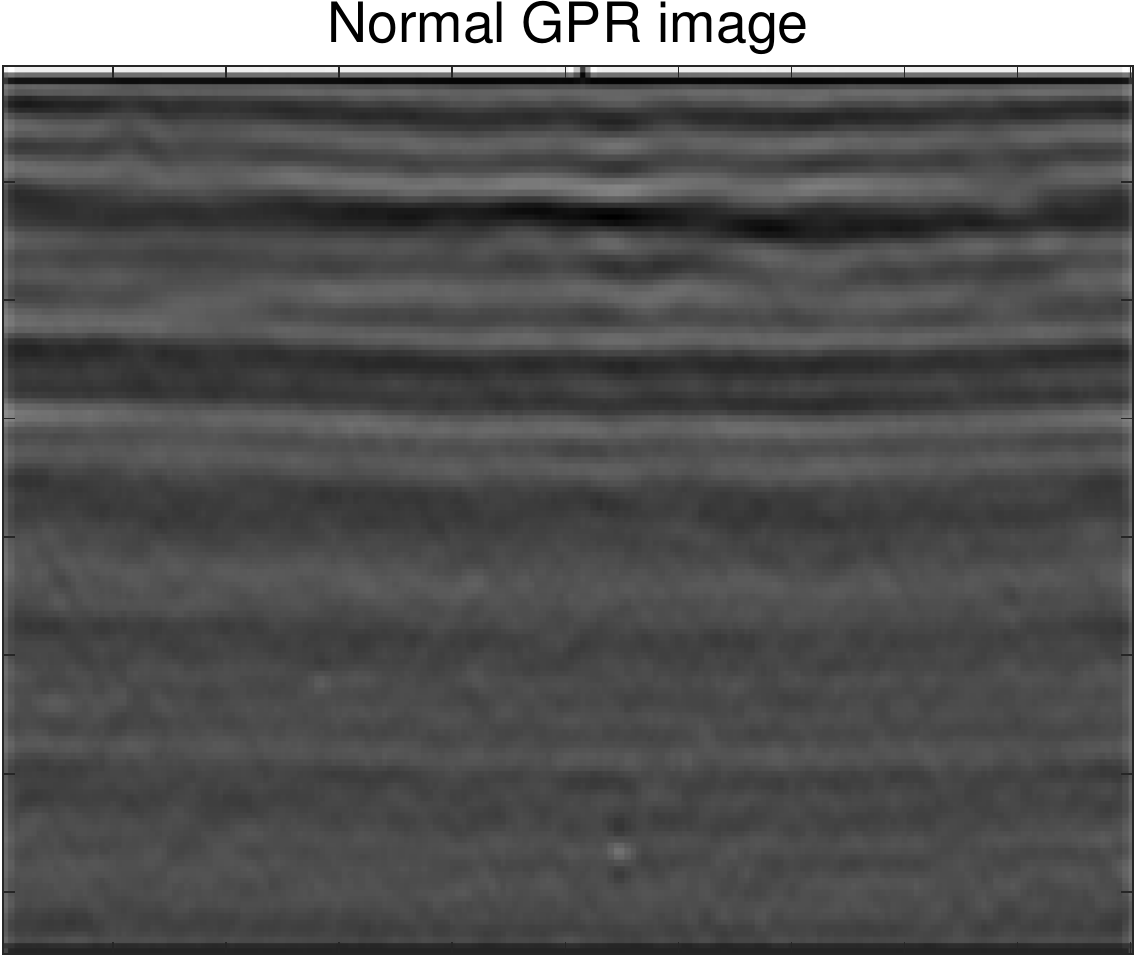}}
	\subfigure[]{ \centering
		\label{n4}
		\includegraphics[width=0.154\textwidth]{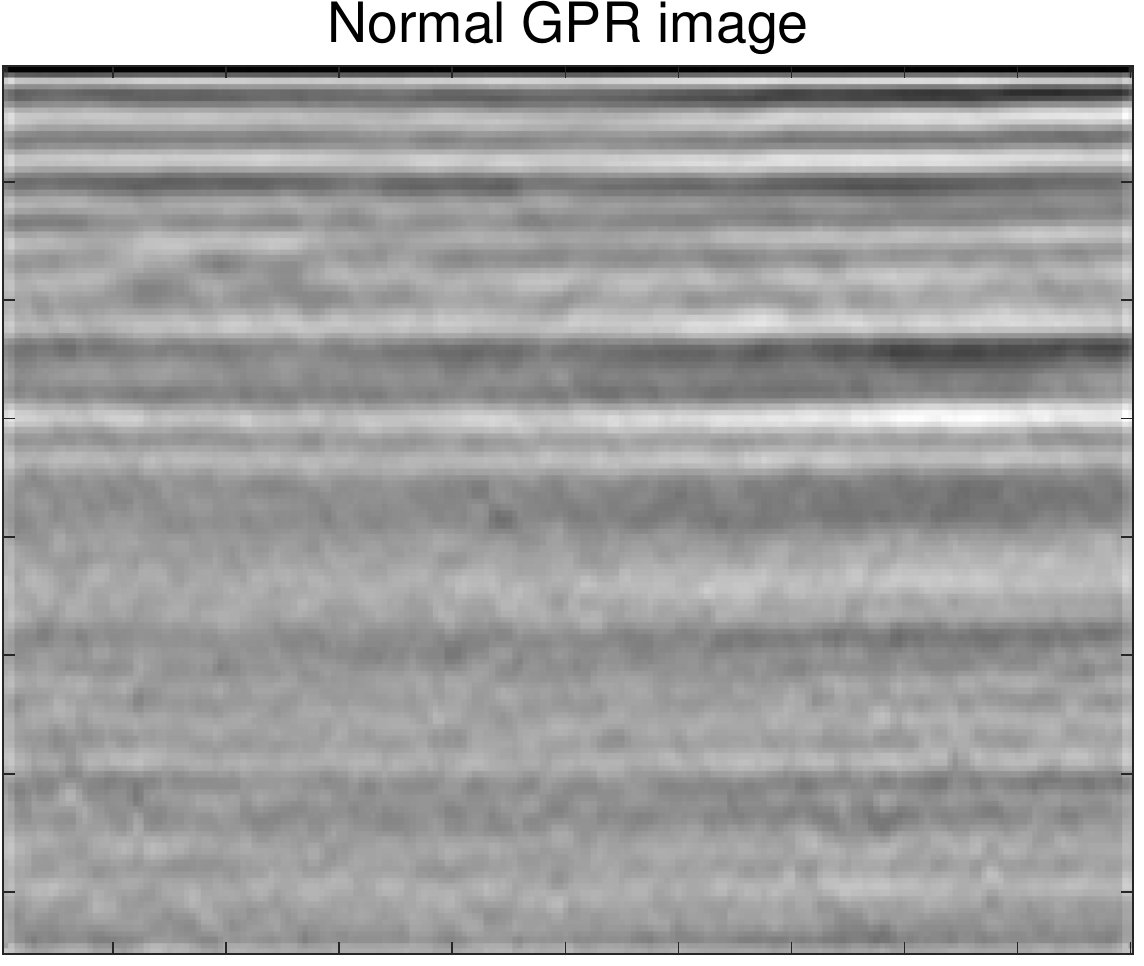}}
	\subfigure[]{ \centering
		\label{n6}
		\includegraphics[width=0.154\textwidth]{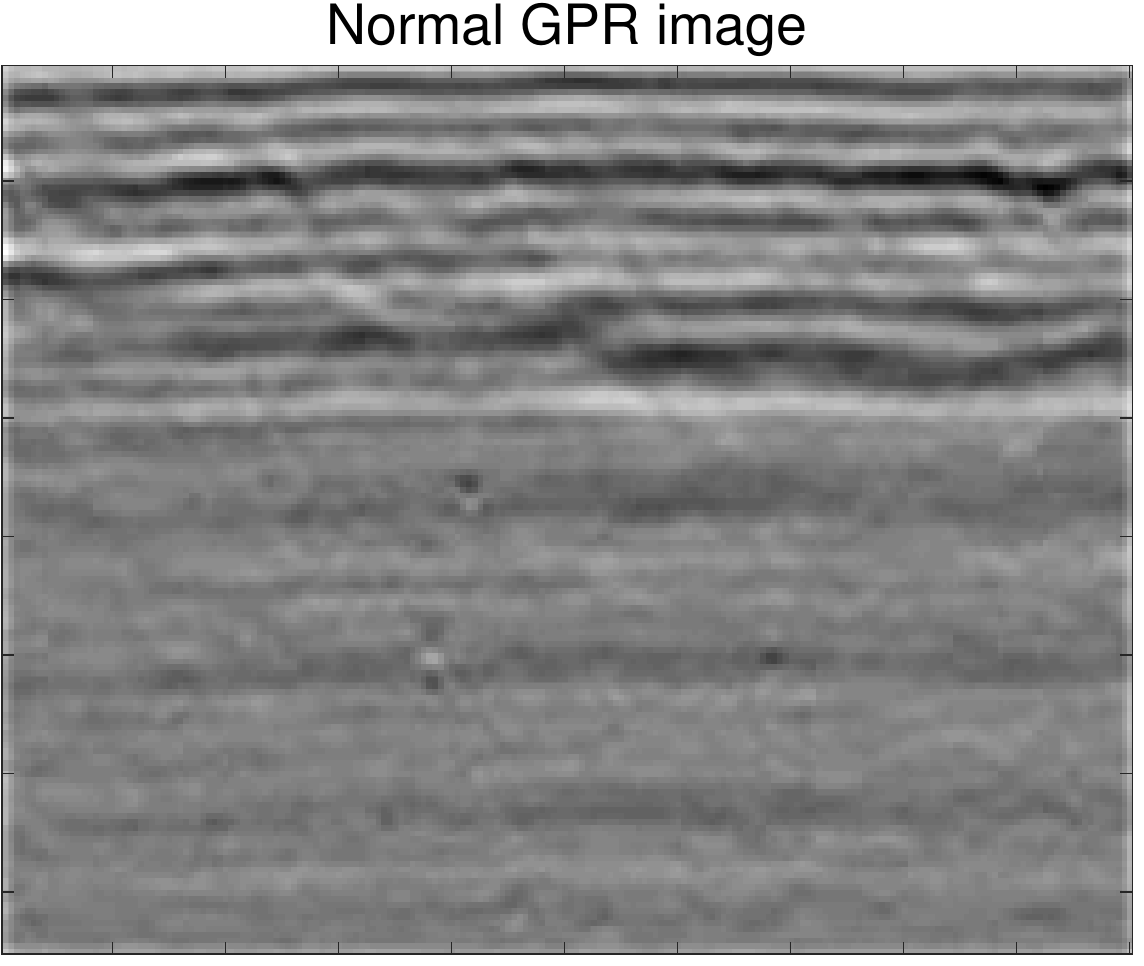}}
%%%
	\subfigure[]{ \centering
		\label{s1}
		\includegraphics[width=0.154\textwidth]{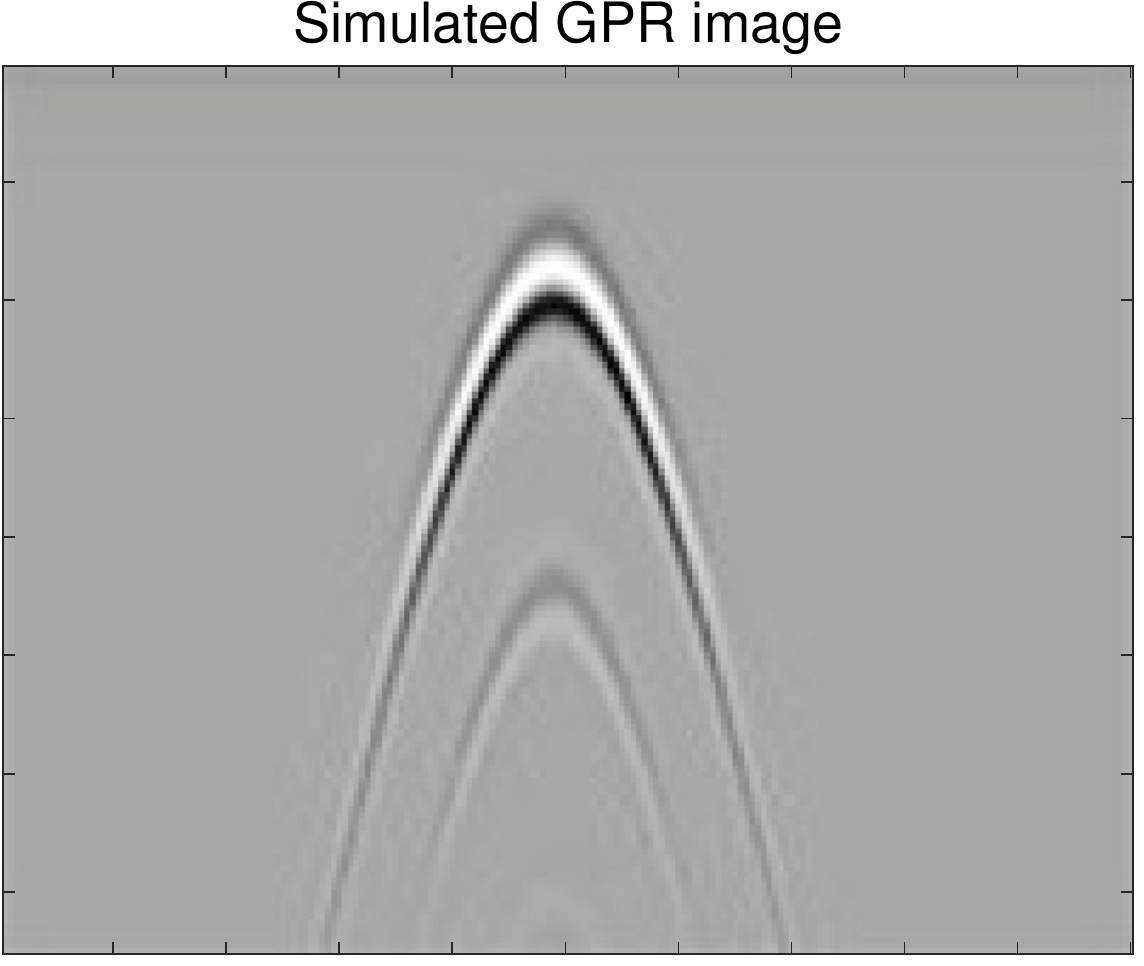}}
	\subfigure[]{ \centering
		\label{s5}
		\includegraphics[width=0.154\textwidth]{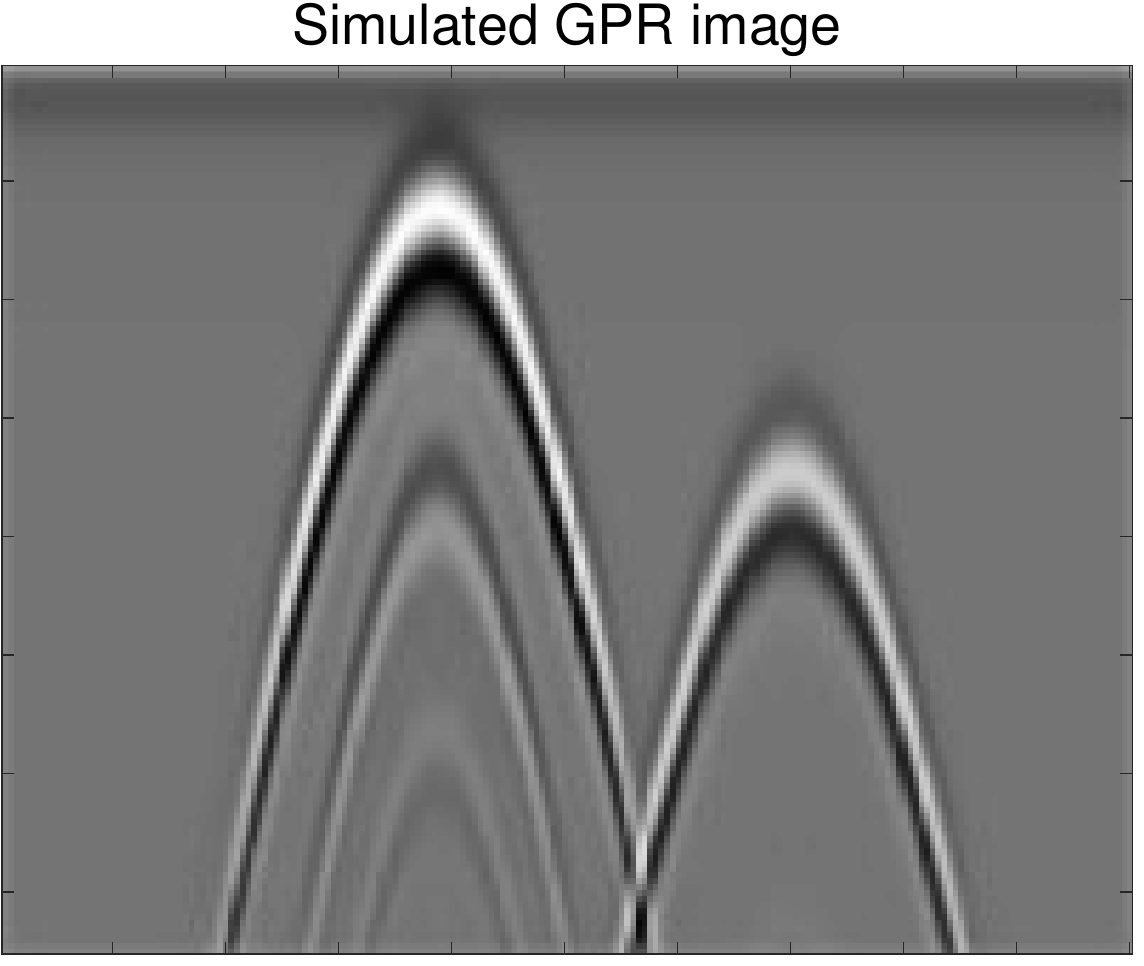}}
	\subfigure[]{ \centering
		\label{s2}
		\includegraphics[width=0.154\textwidth]{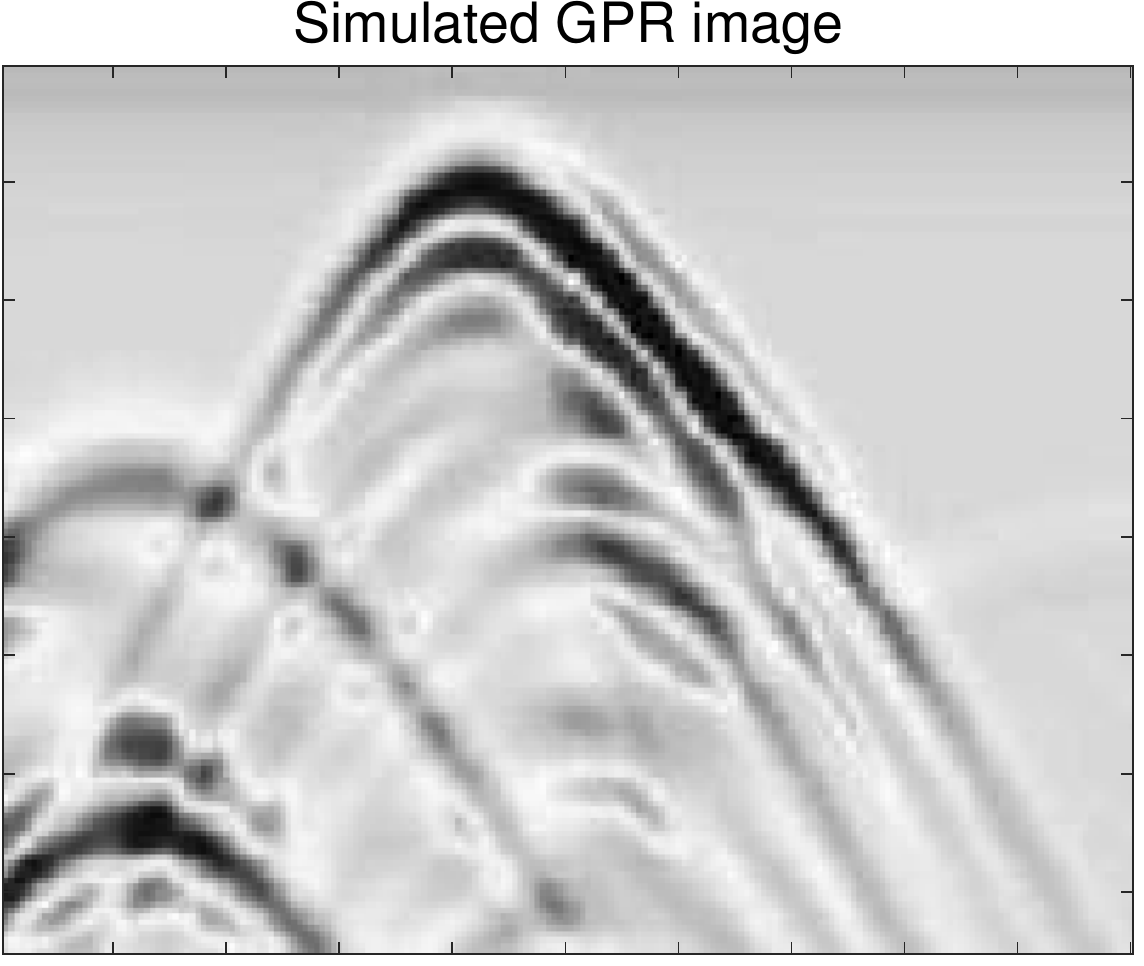}}
	\subfigure[]{ \centering
		\label{s3}
		\includegraphics[width=0.154\textwidth]{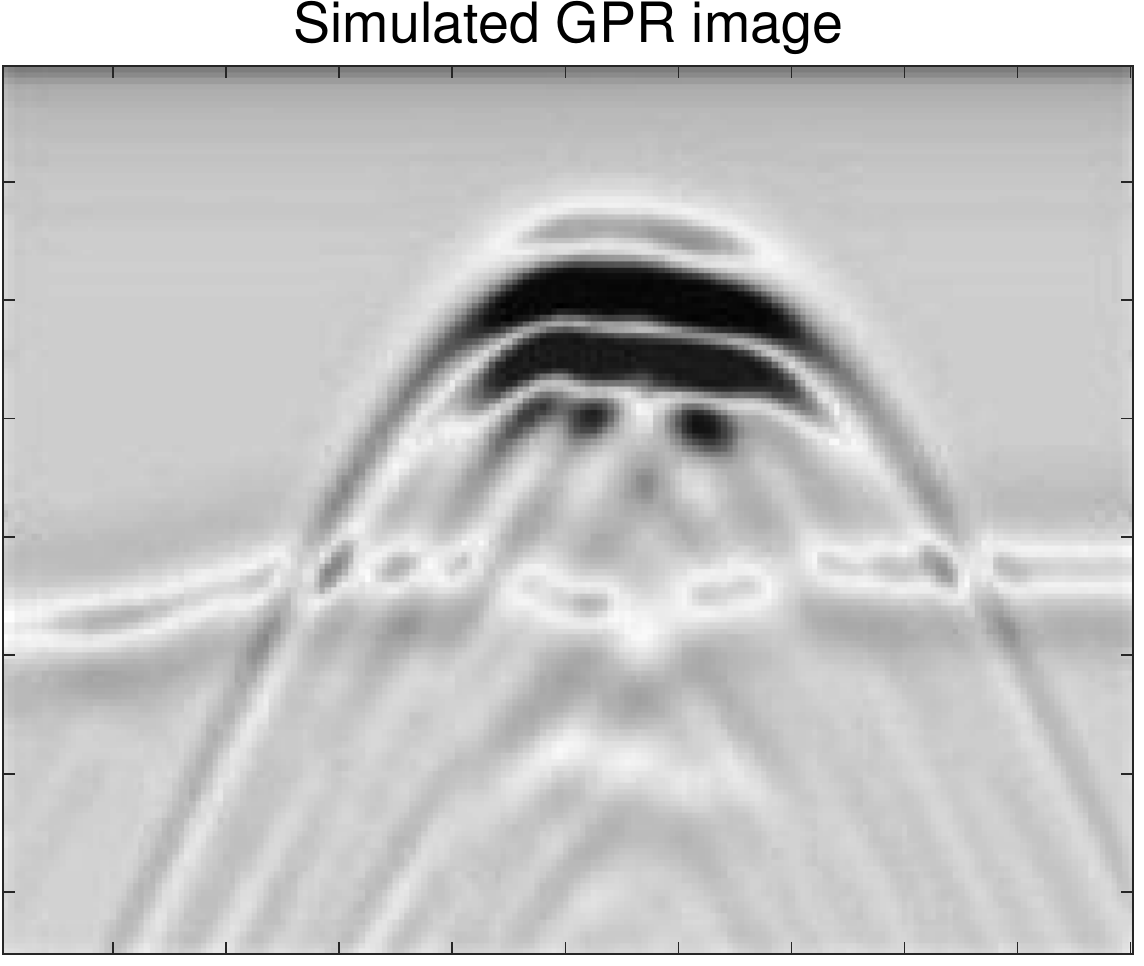}}
	\subfigure[]{ \centering
		\label{s4}
		\includegraphics[width=0.154\textwidth]{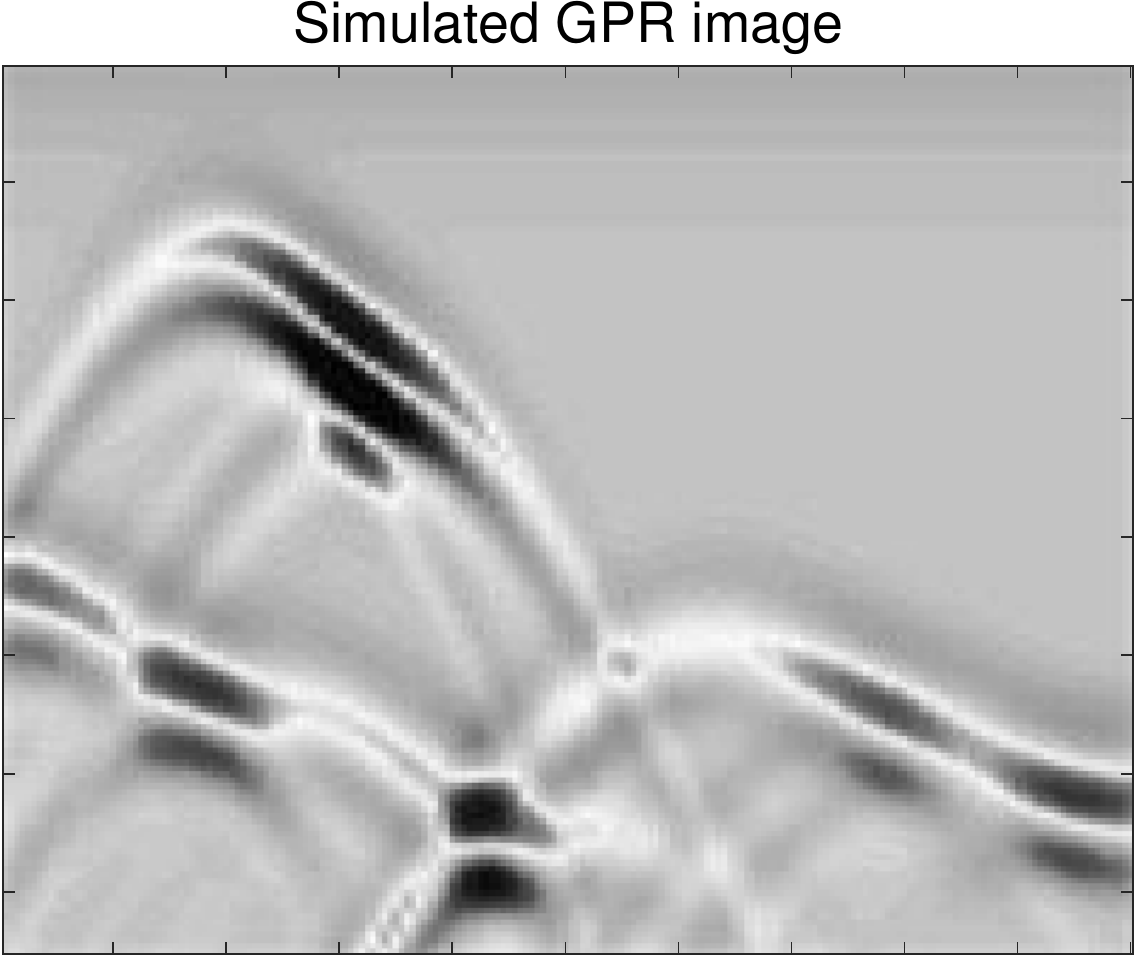}}
	\subfigure[]{ \centering
		\label{s6}
		\includegraphics[width=0.154\textwidth]{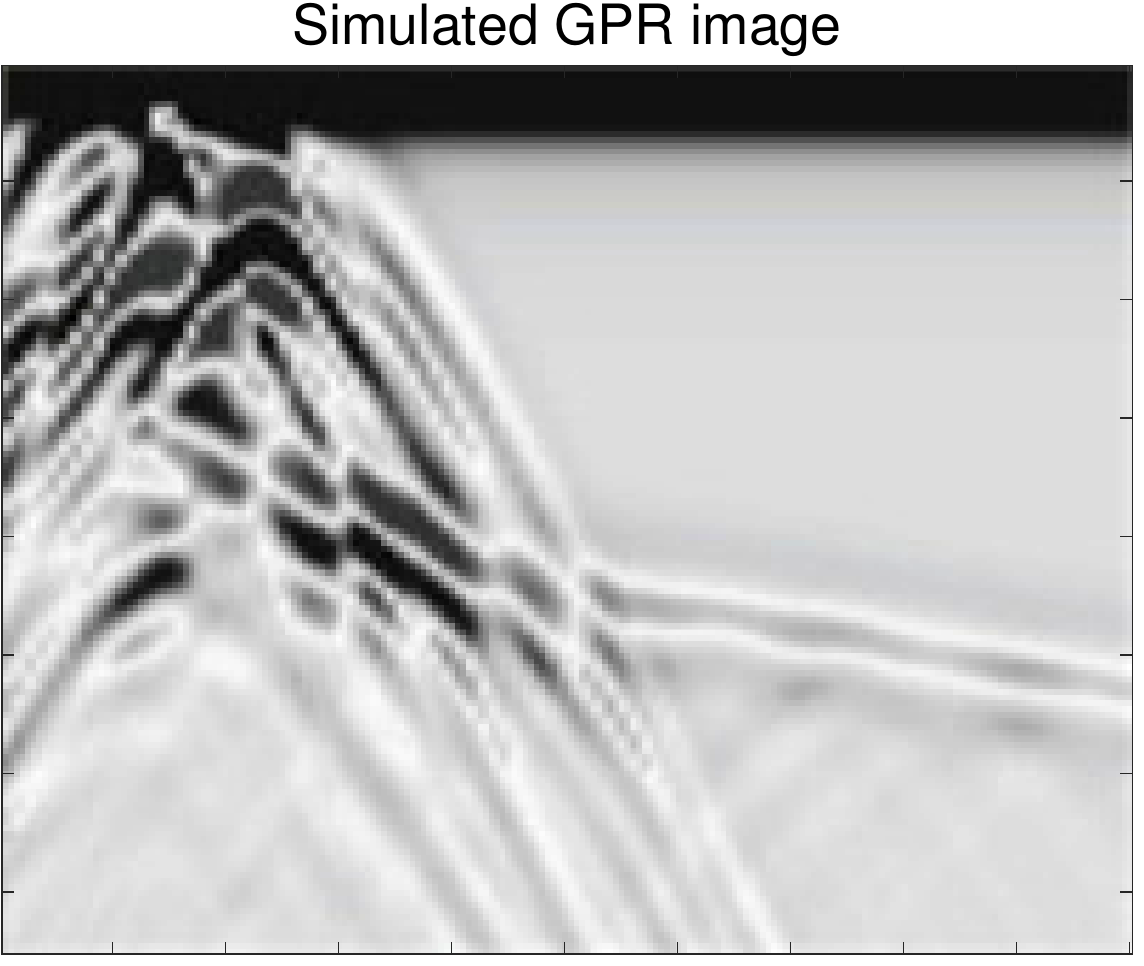}}
%%%
	\subfigure[]{ \centering
		\label{g1}
		\includegraphics[width=0.154\textwidth]{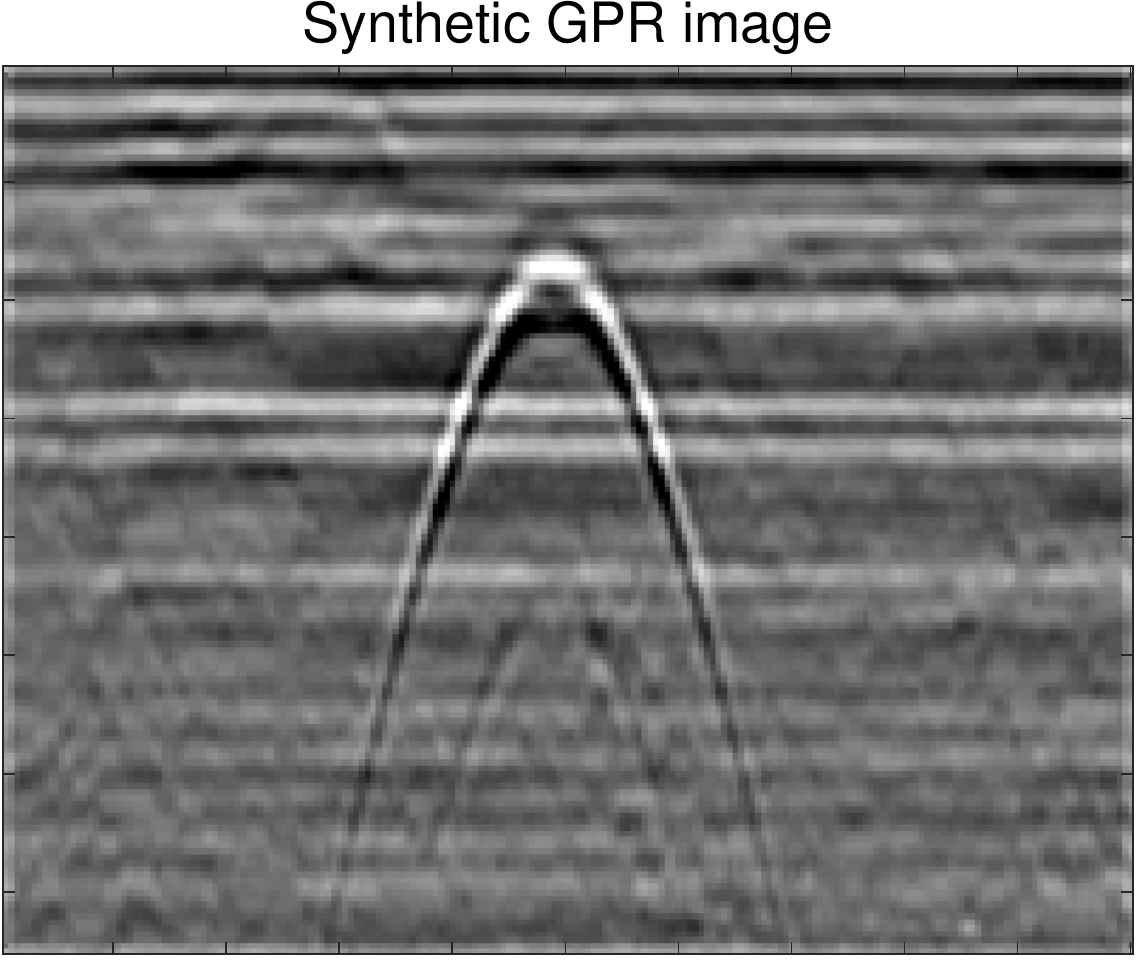}}
	\subfigure[]{ \centering
		\label{g5}
		\includegraphics[width=0.154\textwidth]{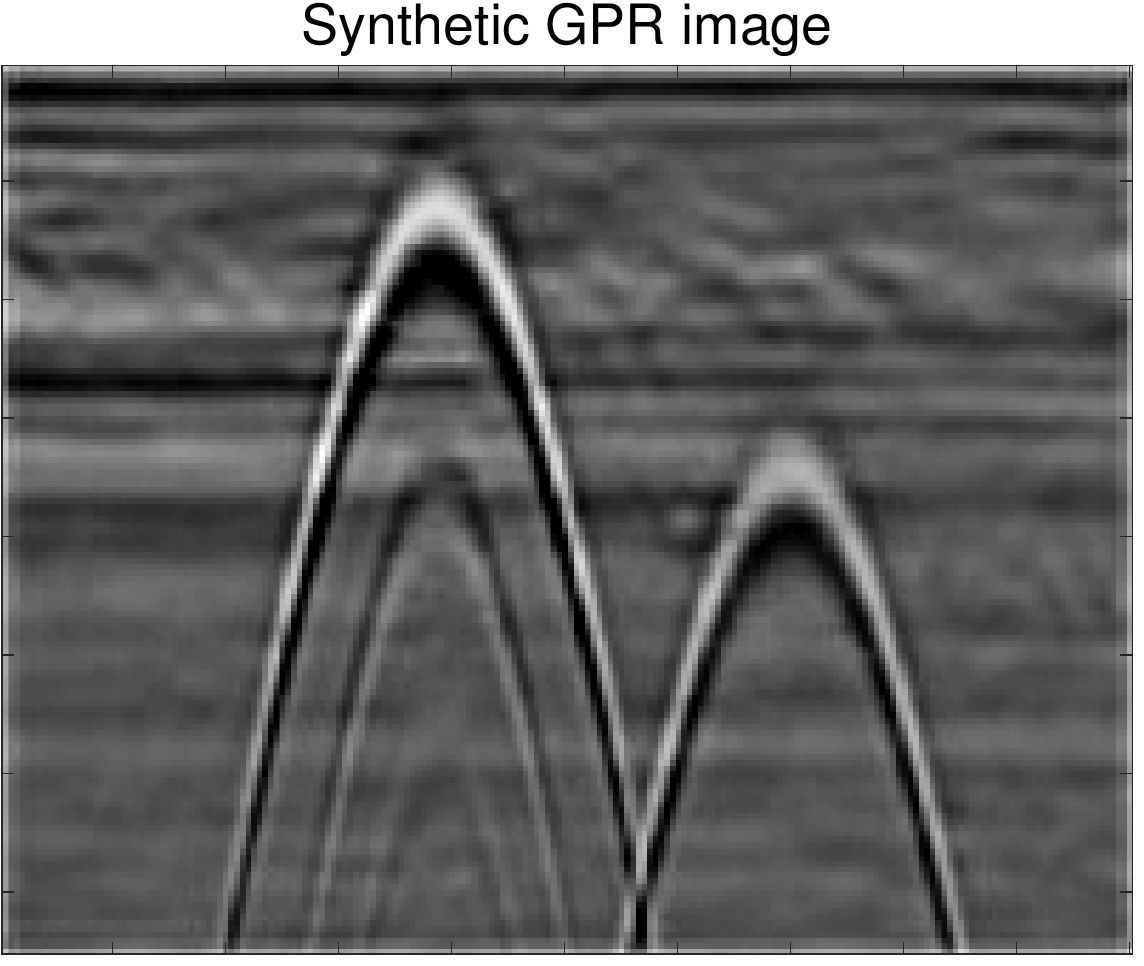}}
	\subfigure[]{ \centering
		\label{g2}
		\includegraphics[width=0.154\textwidth]{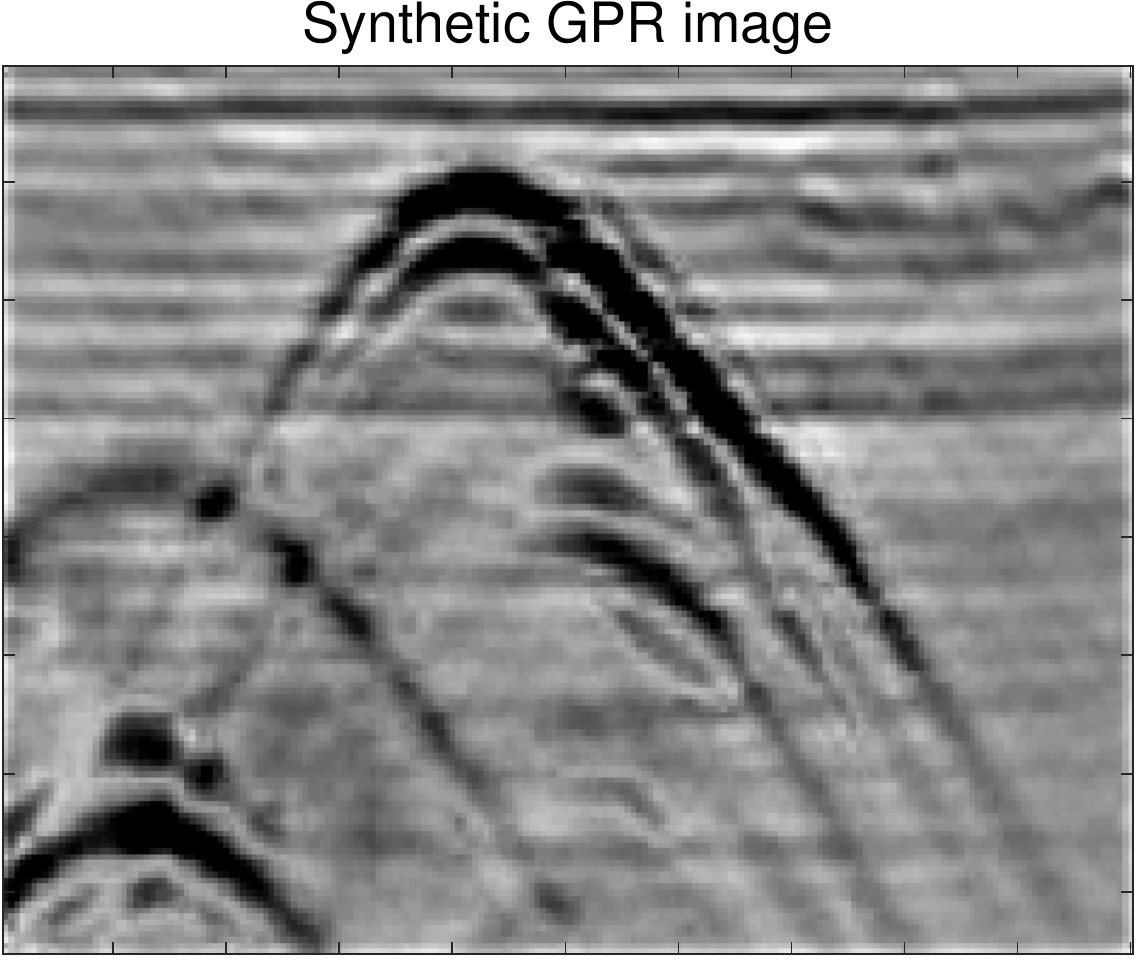}}
	\subfigure[]{ \centering
		\label{g3}
		\includegraphics[width=0.154\textwidth]{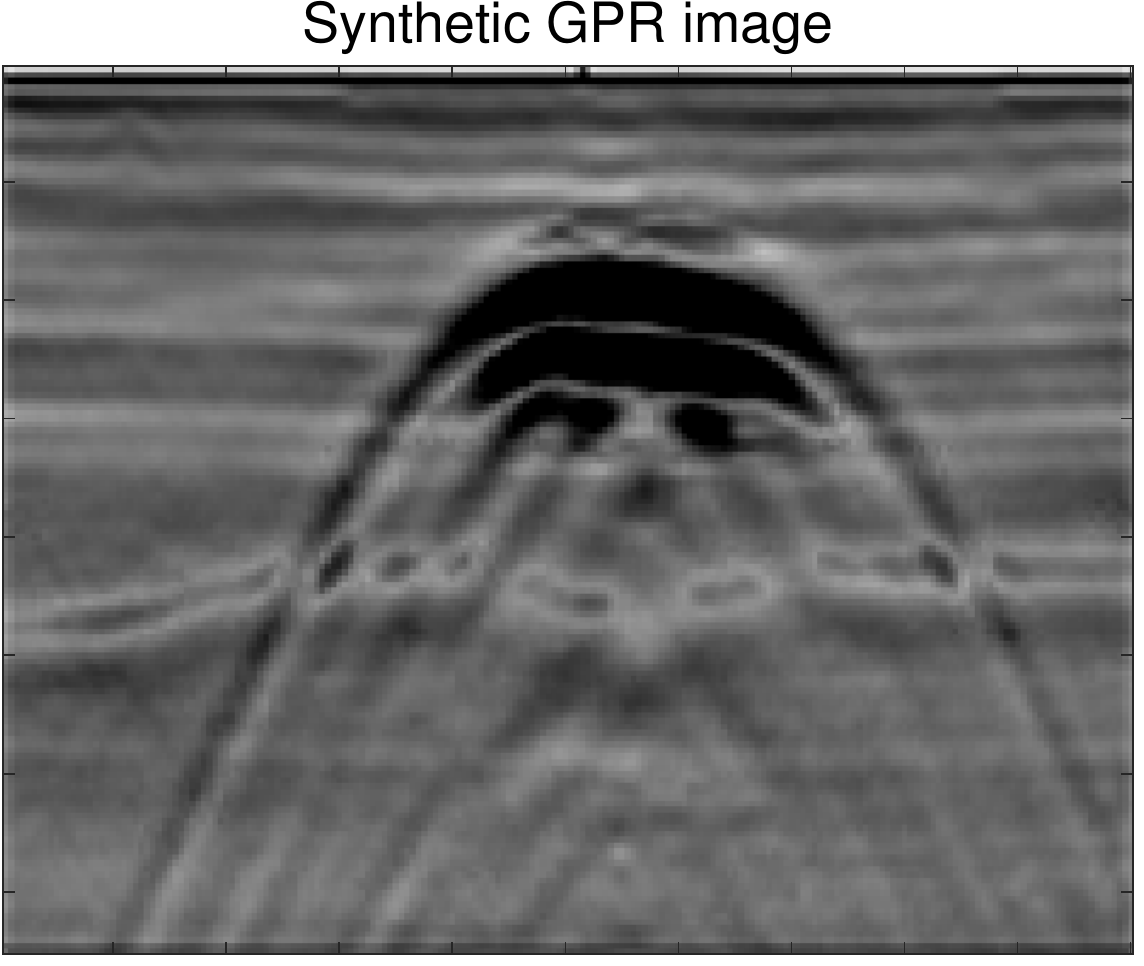}}
	\subfigure[]{ \centering
		\label{g4}
		\includegraphics[width=0.154\textwidth]{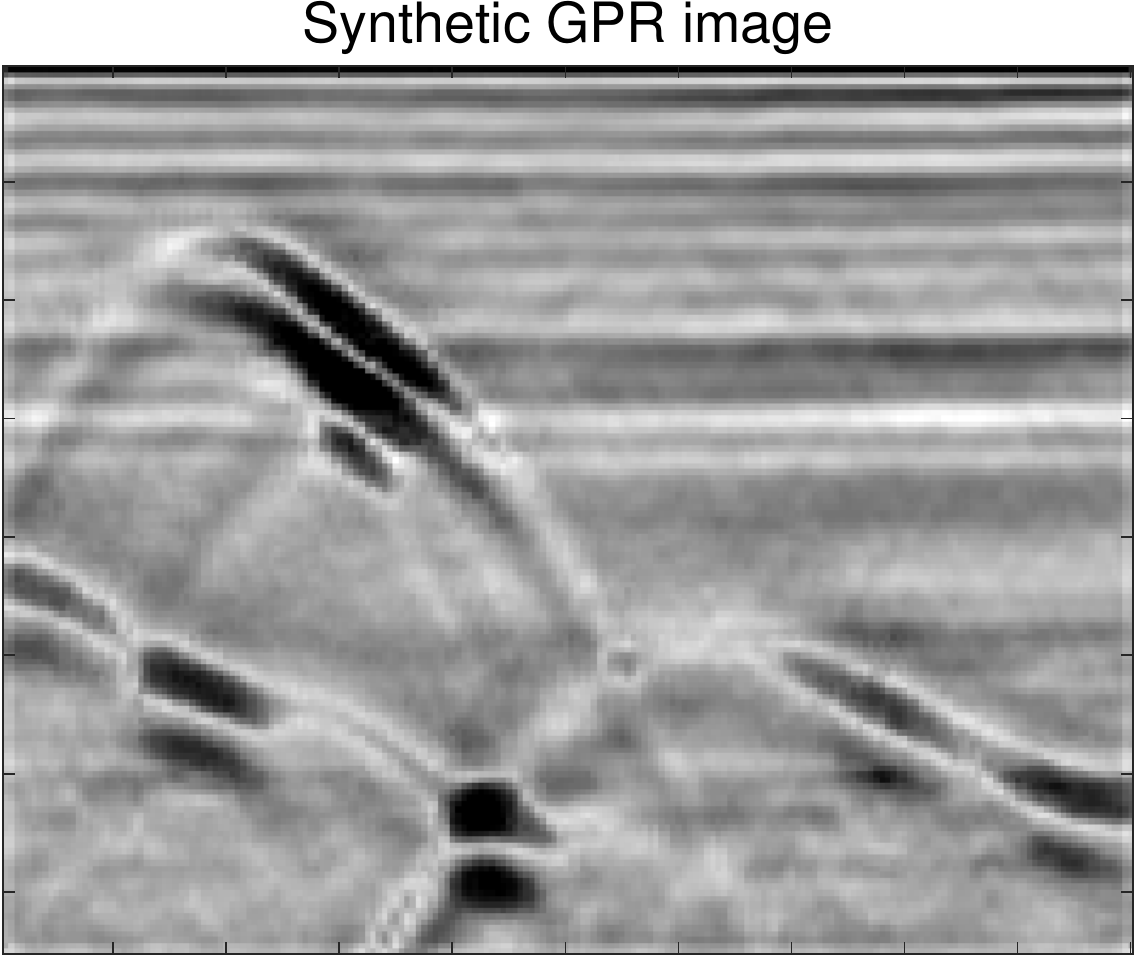}}
	\subfigure[]{ \centering
		\label{g6}
		\includegraphics[width=0.154\textwidth]{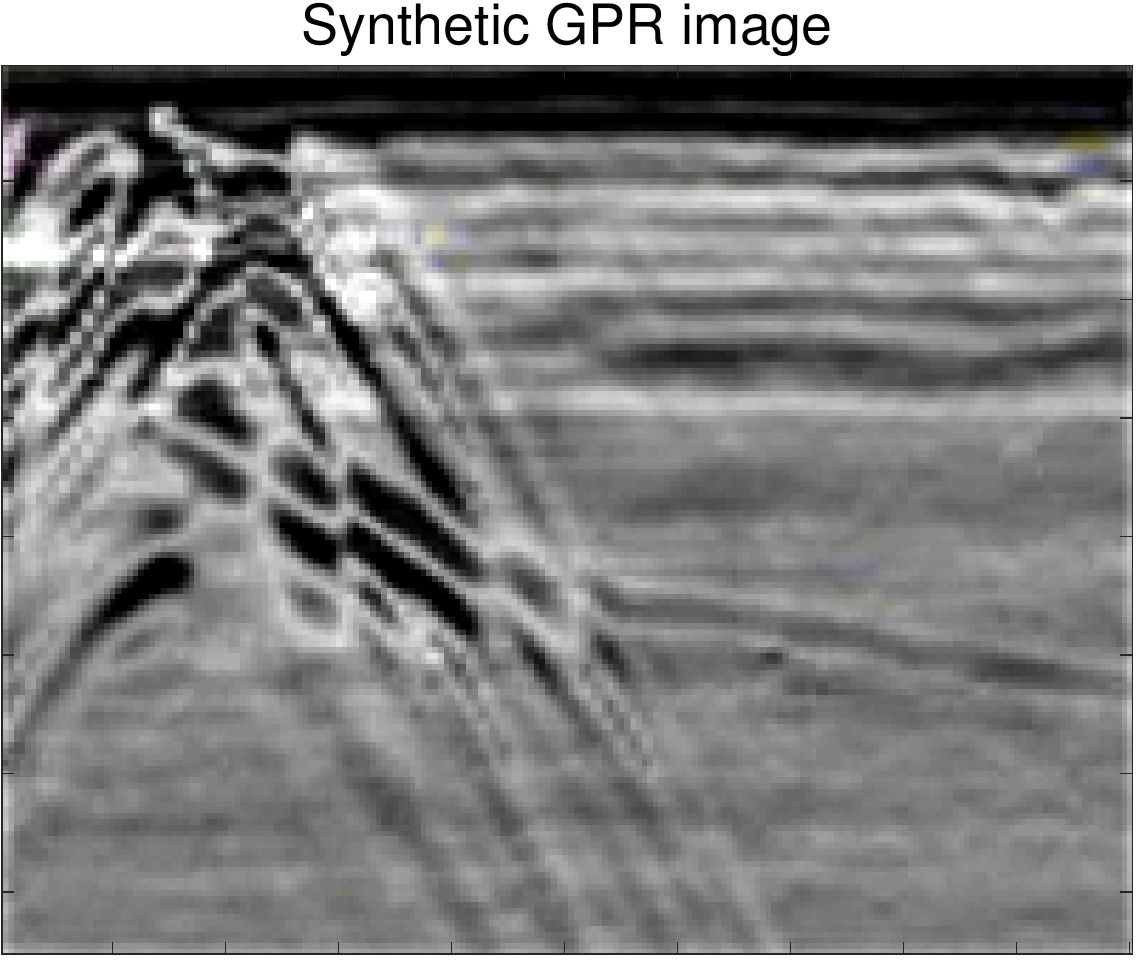}}
%%%
	\caption{(a) to (f) show some normal GPR image segments of the detection areas with no underground objects. (g) to (l) show the simulated GPR image with underground objects generated by GprMax. Specifically, (g) and (h) with hyperbolic shapes are generated by underground metal pipelines. (i) is generated by tunnel lining with water-bearing defects. (j) is generated by a simulated crack and a under-lyed rock. (k) and (l) are generated by a crack and a void. (m) to (r) are the synthetic images generated from the upper two rows of normal and simulated GPR images, each of which is fused by the top two vertically. It could be observed that the synthesized image contains both the basic underground conditions of the detection area as (a) to (f) and the features generated by the underground structures shown in (e) to (h). }
	\label{syndata}
\end{figure*}

The simulated GPR images with different cases of underground structures are generated by GprMax, which is an open-source software, developed on the basis of the Finite-Difference Time-Domain (FDTD) numerical method. GprMax discretizes Maxwell’s equations in both space and time and obtains an approximated solution directly in the time domain through an iterative process. With GprMax, a total of 200 simulated GPR images are generated. The simulated underground objects include: underground cracks, cavities, void, water-bearing, and buried pipes. Part of the settings about the simulated material are presented in Table \ref{setting1}. 
\begin{table}[htbp]
	\small
	\setlength{\tabcolsep}{0pt}
	\renewcommand{\arraystretch}{0.59}
	\centering
	\caption{The Settings of Materials in the Simulated Data Generated by GprMax}
	\begin{tabular}{ccc}\toprule
		Media & \makecell[c]{Relative\\Permittivity} & Details \\\midrule
		Air   & 1     & \makecell[c]{Underground cavities, voids,\\ and above-ground environments} \\\midrule
		Water & 81    & Water-bearing structures \\\midrule
		Rocks & 5 to 8 & \makecell[c]{Subsurface rocks of\\ different shapes and sizes} \\\midrule
		Asphalt & 3 to 5 & Underground environment of asphalt road \\\midrule
		Cement & 4 to 6 & Underground environment of cement road \\\midrule
		Soil (moist) & 8 to 14 & Dry subterranean soil environment \\\midrule
		Soil (dry) & 3     & Moist subterranean soil environment \\\midrule
		\makecell[c]{Conductive\\metal} & $\infty$ & Metal pipes, lines, and cables \\\bottomrule
	\end{tabular}%
	\label{setting1}%
\end{table}%
Each kind of underground objects is set in a variety of sizes, shapes and depths. Taking the pipeline as an example, we set up different forms such as single pipeline, multiple parallel pipelines, and multiple intersecting pipelines. Due to the space limitation of the paper, the details of these objects could not be fully presented, and some cases are shown in Fig. \ref{syndata}.

\subsubsection{The Image Fusion Method}

\begin{figure}[htpb]
	\centering
	\includegraphics[width=0.47\textwidth]{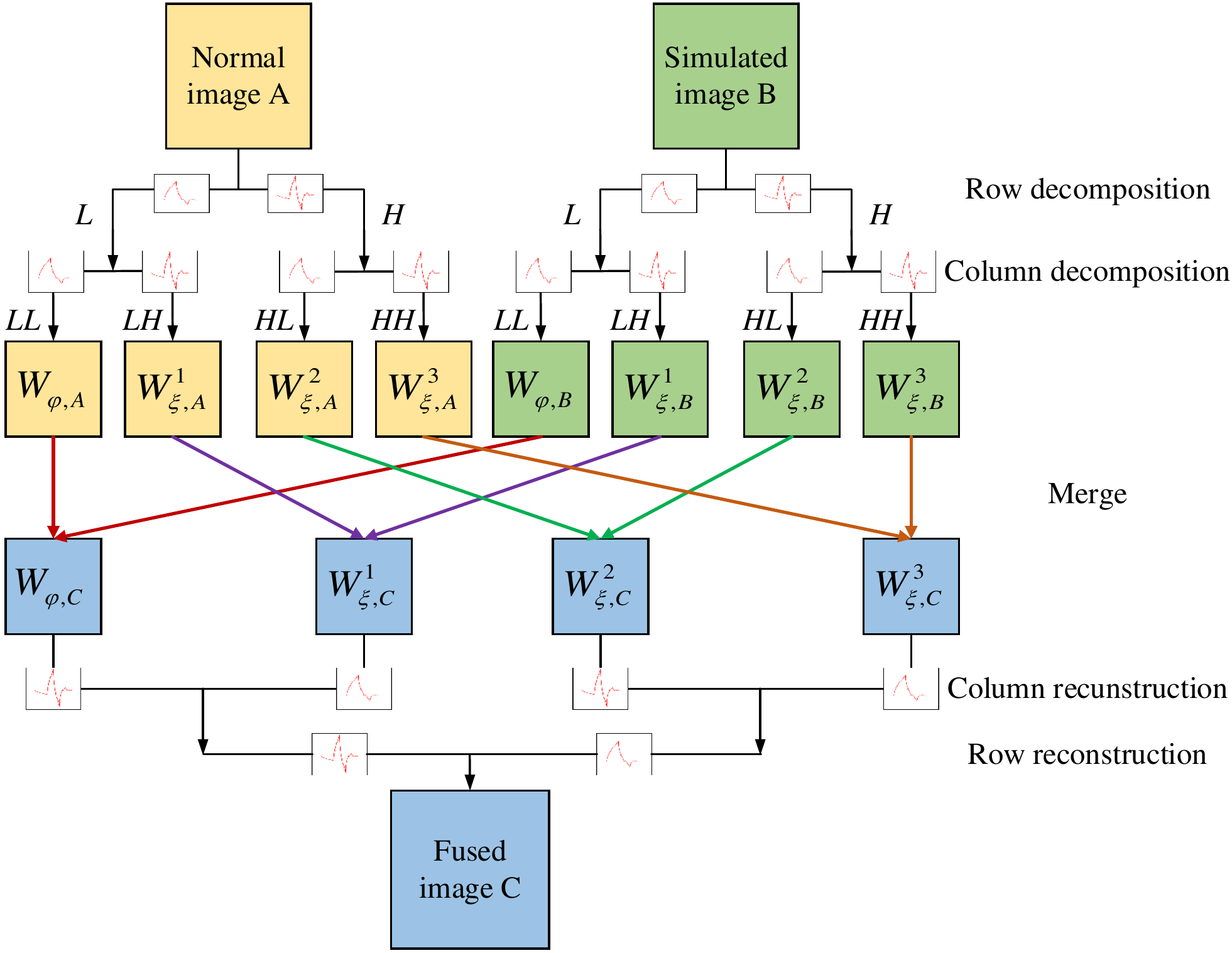}
	\caption{The procedure of the GPR image fusion. A normal GPR image and a simulated GPR image are firstly decomposed into low-frequency component $L$ and high-frequency component $H$ by rows using the scaling function and wavelet function of Daubechies-2 wavelet\cite{1988Orthonormal}. Then, these components are decomposed into approximation decompositions (i.e., $LL$) and detail decompositions (i.e., $LH$, $HL$, $HH$) by columns. After merging these decompositions respectively, the fused image is generated by the wavelet reconstruction method, which is the reverse process of the previous.}
	\label{wavelet}
\end{figure}

Once the normal GPR images and simulated GPR images are obtained, the following work is to generate synthetic images by fusing the two kinds of images. Each fusion takes one image from the normal GPR image segment and one simulated GPR image with underground structures, and one fused image is synthesized. Considering that the GPR image is essentially a representation of wave intensity and propagation time as Fig. \ref{abscan}, and to preserve both the subsurface background and target details, the wavelet decompositions of the two original images are merged to generate the synthetic image as shown in Fig. \ref{wavelet}. Specific operations are presented as follows.

Supposing a segmented normal GPR image $A$ and a simulated image $B$, the approximation decomposition coefficients $W_{\varphi,A}$ and detail decomposition coefficients $W_{\psi,A}^{l=1,2,3}$ could be estimated as:
\begin{equation}
W_{\varphi,A}(0,m,n)=\frac{1}{\sqrt{UV}}\sum_{u,v}f_{A}(u,v)\varphi_{0,m,n}(u,v),
\end{equation}
\begin{equation}
W_{\psi,A}^{l}(j,m,n)=\frac{1}{\sqrt{UV}}\sum_{u,v}f_{A}(u,v)\psi_{j,m,n}^{l}(u,v),
\end{equation}
where $U$, $V$ indicates the size of image $A$, $f_{A}(u,v)$ is the value of $A$ at position-$(u,v)$. Moreover, $\varphi_{0,m,n}$ and $\psi_{j,m,n}^{l}$ are wavelet basis functions derived from the scaling function $\varphi$ and wavelet functions $\psi^{l}$ of Daubechies-2 wavelet\cite{1988Orthonormal}:
\begin{equation}
\varphi_{j,m,n}(u,v)=2^{j/2}\varphi(2^{j}u-m,2^{j}v-n),
\end{equation}
\begin{equation}
\psi_{j,m,n}^{l}(u,v)=2^{j/2}\psi^{l}(2^{j}u-m,2^{j}v-n).
\end{equation}

The $W_{\varphi,B}$ and $W_{\psi,B}^{l}$ could be estimated in similar ways, then a wavelet transform array is synthesized and populated by taking the minimum of $W_{\varphi,A}$, $W_{\varphi,B}$ as $W_{\varphi}$ and the maximum of $W_{\psi,A}^{l}$, $W_{\psi,B}^{l}$ as $W_{\psi}^{l}$. As aforementioned, the purpose of the above settings is to simultaneously preserve the basic GPR image characteristics of the underground environment and the detailed characteristics of the simulated underground objects. After that, the inverse wavelet transform could be applied to create the fused image $C$:
\begin{equation}
\begin{aligned} 
f_{C}(u,v) &= \frac{1}{\sqrt{UV}}\sum_{m,n}W_{\varphi}(0,m,n)\varphi_{0,m,n}(u,v)+\\
&\frac{1}{\sqrt{UV}}\sum_{j=0}^{\infty}\sum_{l,m,n}W_{\psi}^{l}(j,m,n)\psi_{j,m,n}^{l}(u,v).
\end{aligned}
\end{equation}

Fig. \ref{syndata} illustrates some segmented normal GPR images, simulated GPR images with underground objects, and fused synthetic images. It could be intuitively observed that the synthesized image contains both the basic underground conditions of the detection area and the features generated by the underground structures simulated in GprMax. Fine-tuning with synthetic data could increase the network's sensitivity to subsurface anomalies in the specific area. In the conducted experiments, we evaluate the feature extraction effect of CNN not fine-tuned or fine-tuned only on normal and simulated data. The obtained results show that fine-tuning by synthetic data could improve both the precision and recall of identifying different underground anomalies in the feature space.

\subsection{Fine-tuning with Synthetic and Normal Data}

Any pre-trained CNN could be used as the feature extractor, while the existing study\cite{oza2018one} on CNN-based classification has shown that training or fine-tuning a pre-trained CNN with specific data could improve the feature extraction effect on this kind of data. 
The above data generation provides the training data of the detection area, which could be divided into two classes:
\begin{enumerate}
	\item The GPR image segment that do not contain underground objects in the detection area (e.g. Fig. \ref{n1} to \ref{n6});
	\item The synthetic GPR image with underground objects (e.g. Fig. \ref{g1} to \ref{g6}).
\end{enumerate} 

To enrich the data used to fine-tune the CNN, and improve the robustness of the adopted network, the data augmentation is performed on the normal and synthetic data. Since the GPR image is essentially a combination and representation of EM waves, and there are physical meanings with the grayscales and directions in the GPR B-scan image\cite{zhou2018automatic}, it could be necessary to keep such properties stable during the data enhancement. For normal GPR images without subterranean objects, the Noise injection \cite{moreno2018forward} and Image mixing \cite{inoue2018data} are utilized. For the synthetic images, the simulated images are scaled to different sizes and then fused with different positions of the normal images. 

Considering the timeliness requirement, the ResNet-18 \cite{he2016deep},  VGG16 \cite{liu2015very} and AlexNet\cite{krizhevsky2012imagenet} that have been pre-trained on ImageNet \cite{deng2009imagenet} are adopted in this paper. It is assumed that the utilized networks could extract the feature vector with the size of $D$.
Since the goal of fine-tuning is to improve the feature extraction of pre-trained CNNs for the GPR image obtained in the specific detection area, rather than directly improve its end-to-end classification efficiency, a full-connect layer with size of $D \times 2$ and the softmax regression layer are connected to the feature vector\cite{oza2018one}, and the following binary cross-entropy loss function is used to train the network:
\begin{equation}
	L=-\frac{1}{N_s}\sum_{j=1}^{N_s}\left( y \log(p)+(1-y)\log(1-p)\right),
\end{equation}
where $y\in\left\{ 0,1 \right\}$, $p$ denotes the softmax probability, and $N_s$ is determined by the batch size.
The network is optimized using the Stochastic Gradient Descent (SGD) \cite{amari1993backpropagation} with
learning rate of $10^{-3}$. The input image batch size of 16 is used
in our approach. In the conducted experiments, all fine-tuning converged within 10 rounds of training, ensuring the timeliness of the method proposed in this paper.

\subsection{Extracting Features with the Fine-tuned CNN}

After the above fine-tuning, as Fig. \ref{Procedure} shows, the fine-tuned CNN is utilized as a feature extractor by removing the above $D \times 2$ full-connect layer and softmax regression layer. A sliding window is then constructed and swiped across the  GPR image subsequently obtained in the detected area. The feature of the GPR image in each sliding window is extracted by the fine-tuned CNN, thus the GPR images obtained in the detection area are mapped in to the feature space.

\section{One-Class learning in the feature space}

According to the unknown numbers and types of subsurface anomalies that may arise, after mapping the GPR B-scan image into the feature space, the one-class learning algorithms are utilized to classify the corresponding underground objects in the feature space. 

The Support Vector Data Description (SVDD) \cite{scholkopf2001estimating} is firstly used to form a hypersphere defined by the radius $R$ and center $\mathbf{a}$ by constructing the hyperboundary, and making it contain as many normal features as possible, so as to achieve the maximum separation of normal and abnormal features.
Suppose the fine-tuned CNN extracts features from normal GPR images within $n$ sliding windows as:
\begin{equation}
	\mathbf{X}=\left\{ \mathbf{x}_1, \mathbf{x}_1, \cdots, \mathbf{x}_n   \right\} \in \mathbb{R}^{n\times d},
\end{equation}
where $n$ is the number of extracted features and $d$ is the feature dimension.
The solution of the hyperplane's boundary could be formed as an optimization problem:
\begin{equation}
	\label{fsvdd}
	\begin{aligned}
		&\text{min}\  F (R, \mathbf{a})= R^2 + C\sum_{i=1}^{n}\xi_i,  \\
		&\text{s.t.}\   \Vert \varphi(\mathbf{x}_i) - \mathbf{a} \Vert \leq R^2 + \xi_i
	\end{aligned}
\end{equation}
where $\mathbf{a}$ and $R$ indicate the center and radius of the hypersphere, $\xi_i \geq 0$ indicates the slack variables (similar to Support Vector Machine (SVM)\cite{suthaharan2016support}), the parameter $C$ controls the trade-off between the volume and the errors. $\varphi (\cdot)$ in this paper is the function that maps the $\mathbf{x}_i\in \mathbf{X}$ in to the inner product space of Gaussian Kernel\cite{suthaharan2016support}. 
By introducing the Lagrange multiplier $\alpha_i \in \left[0, C\right]$, the above Eq. \eqref{fsvdd} could be transformed into the duality problem as 
\begin{equation}
	\label{fsvdd2}
	\begin{aligned}
		&\text{max}\  L=\sum_{i=1}^{n} \alpha_i \boldsymbol{K} ( \mathbf{x}_i, \mathbf{x}_i)-\sum_{i=1}^{n}\sum_{j=1}^{n} \alpha_i \alpha_j\ \boldsymbol{K} ( \mathbf{x}_i ,\mathbf{x}_j),  \\
		&\text{s.t.}\   \sum_{i=1}^{n}\alpha_i = 1,\  0\leq \alpha_i \leq C,
	\end{aligned}
\end{equation}
where $\boldsymbol{K}$ indicates the inner product function of Gaussian kernel. 
The above Eq. \ref{fsvdd2} could be solved by existing optimization
methods for SVM \cite{tax1999support}, and the hypersphere could be obtained. Thus the initial classifier that could identify the GPR image is normal or abnormal is conducted by testing an extracted feature $\mathbf{x}$ to be detected as an anomaly if $\Vert \varphi(x) - \mathbf{a} \Vert \geq R ^2$.

\begin{figure}[htbp]
	\centering
	\subfigure[]{ \centering
		\label{oneclass}
		\includegraphics[width=0.23\textwidth]{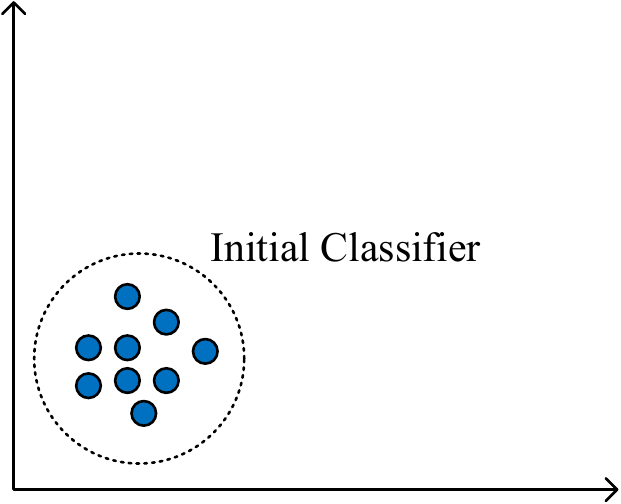}}
	\subfigure[]{ \centering
		\label{oneclass_inc}
		\includegraphics[width=0.23\textwidth]{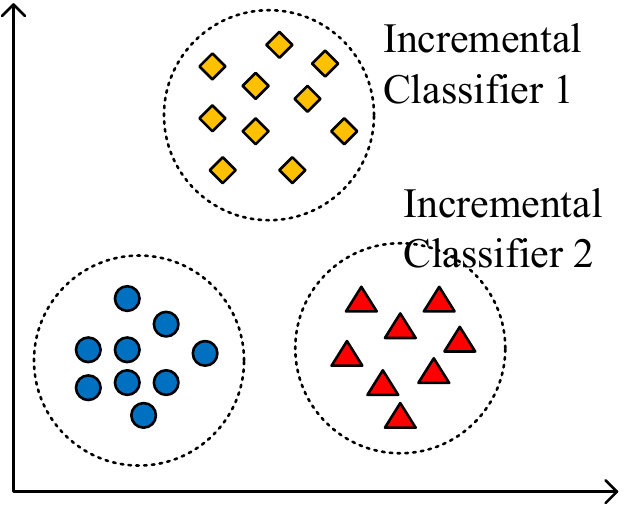}}
	\caption{(a) is the schematic diagram of one-class learning. Blue points are positive data. One-class learning only uses positive data to train a positive-negative classifier. (b) is the schematic diagram of incremental one-class learning. The points in the three circles represent the three types of data. And incremental one-class learning uses each one-class classifier to represent each kind of data.}
	\label{ocleanring}
\end{figure}

In the proposed method, the normal GPR B-scan image segments obtained in the detection area without any underground objects are firstly mapped to the feature space via the fine-tuned CNN, and a initial SVDD classifier is trained. 
Then, for the subsequent image in the sliding window, the features would be extracted and continuously classified by the initial classifier trained from a normal GPR B-scan image segment. 
The image in the window classified as abnormal would be put into the abnormal set. The abnormal data is further trained by incremental One-Class Support Vector Machinev \cite{chen2013learning} (i.e. incremental OCSVM), where more incremental classifiers could be obtained to classify the features into several classes. Thus the corresponding GPR image segments are classified and the underground object that generate the GPR image segment could be identified and classified.
Fig. \ref{ocleanring} illustrates the schematic diagrams of one-class learning and incremental one-class learning. Based on the above operations, the underground anomaly diagnosis could be performed on a detection area without presetting any type and number of anomalies.

\section{Experimental Study}

In this section, experiments on real-world datasets are conducted. After that, the analysis of the experimental results and some comparative works are presented.

\subsection{Experiments on Real-World Datasets}

\begin{figure*}[htpb]
	\centering
	\subfigure[]{ \centering
		\label{data32}
		\includegraphics[height=1.62in]{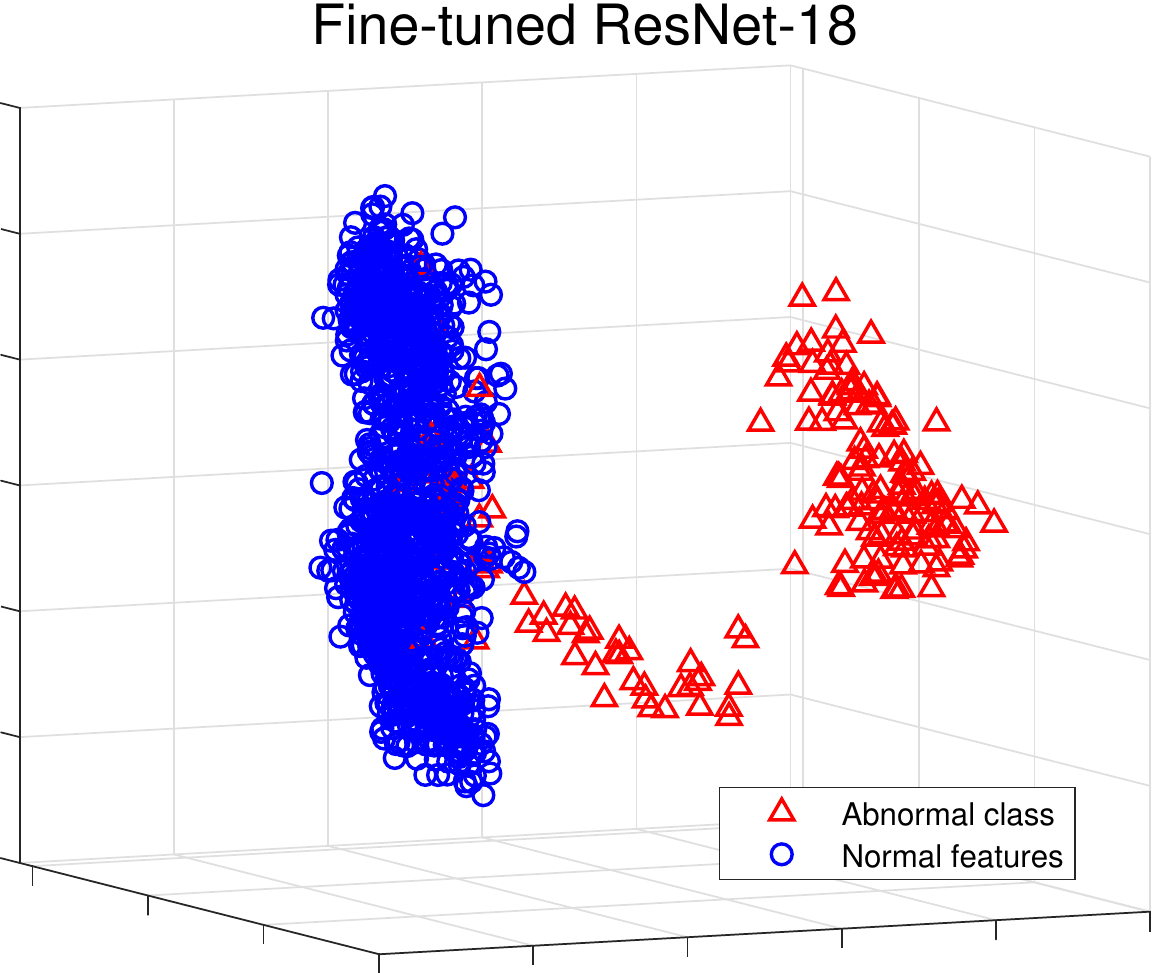}}
	\subfigure[]{ \centering
		\label{data92}
		\includegraphics[height=1.62in]{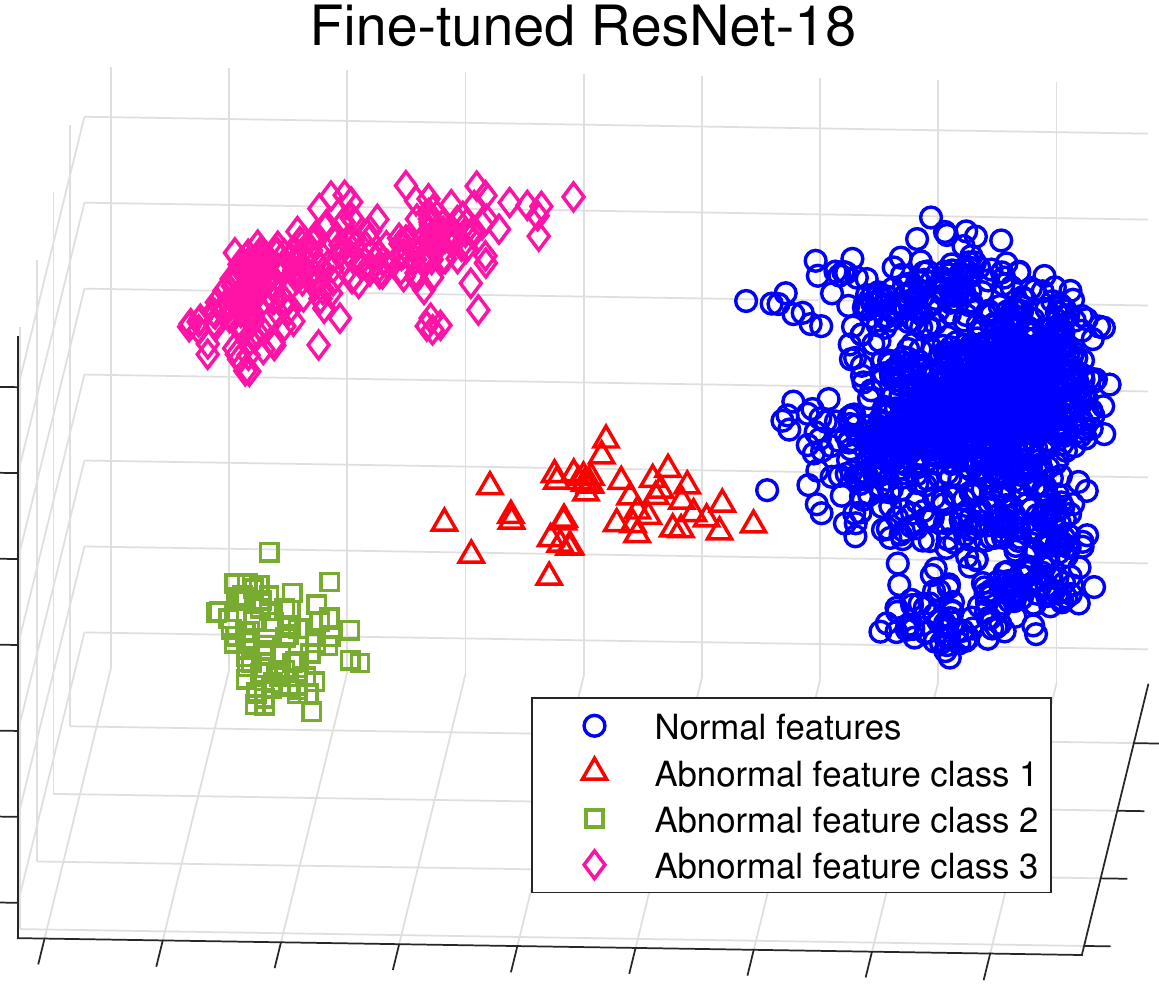}}
	\subfigure[]{ \centering
		\label{data22}
		\includegraphics[height=1.62in]{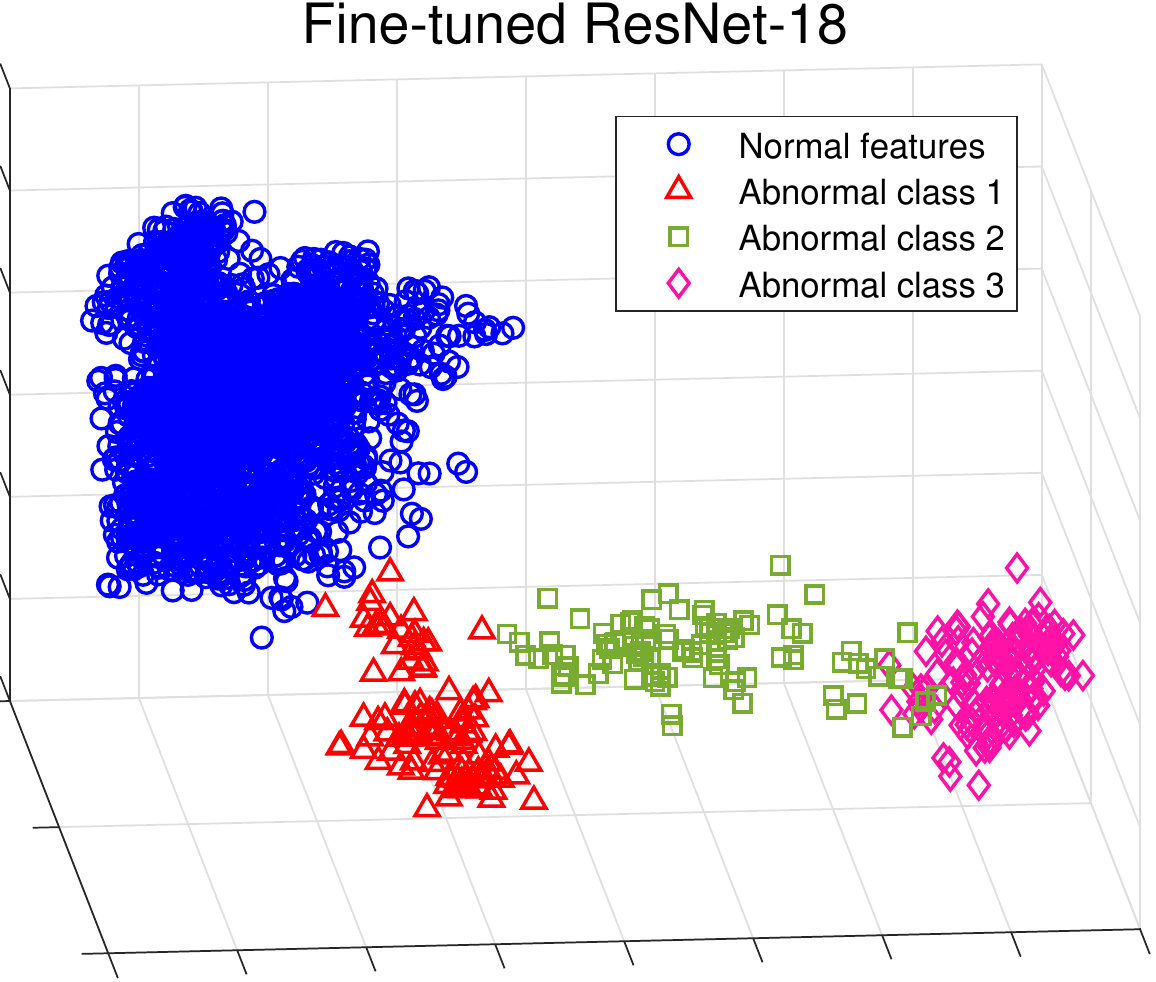}}
	\subfigure[]{ \centering
		\label{data09}
		\includegraphics[height=1.62in]{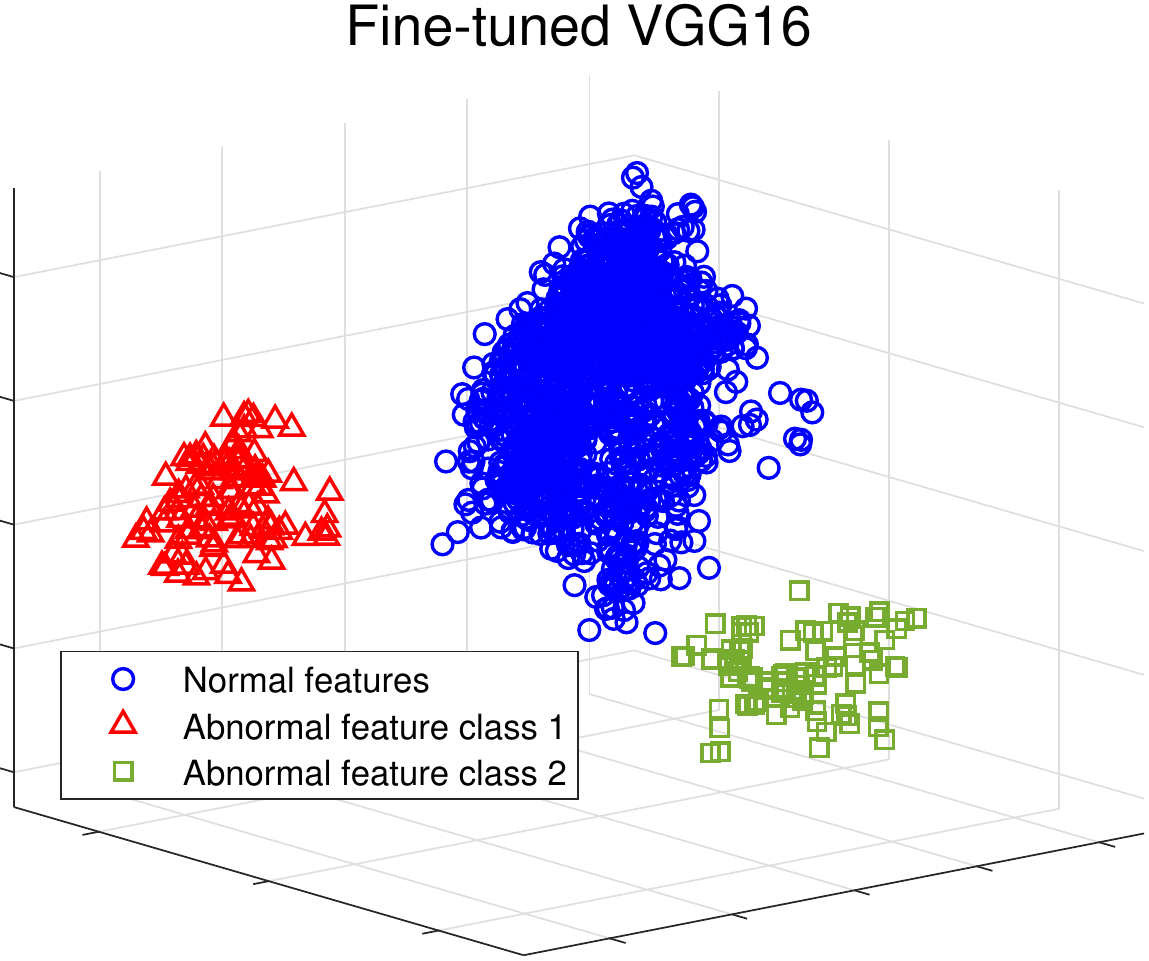}}
	\subfigure[]{ \centering
		\label{data03}
		\includegraphics[height=1.62in]{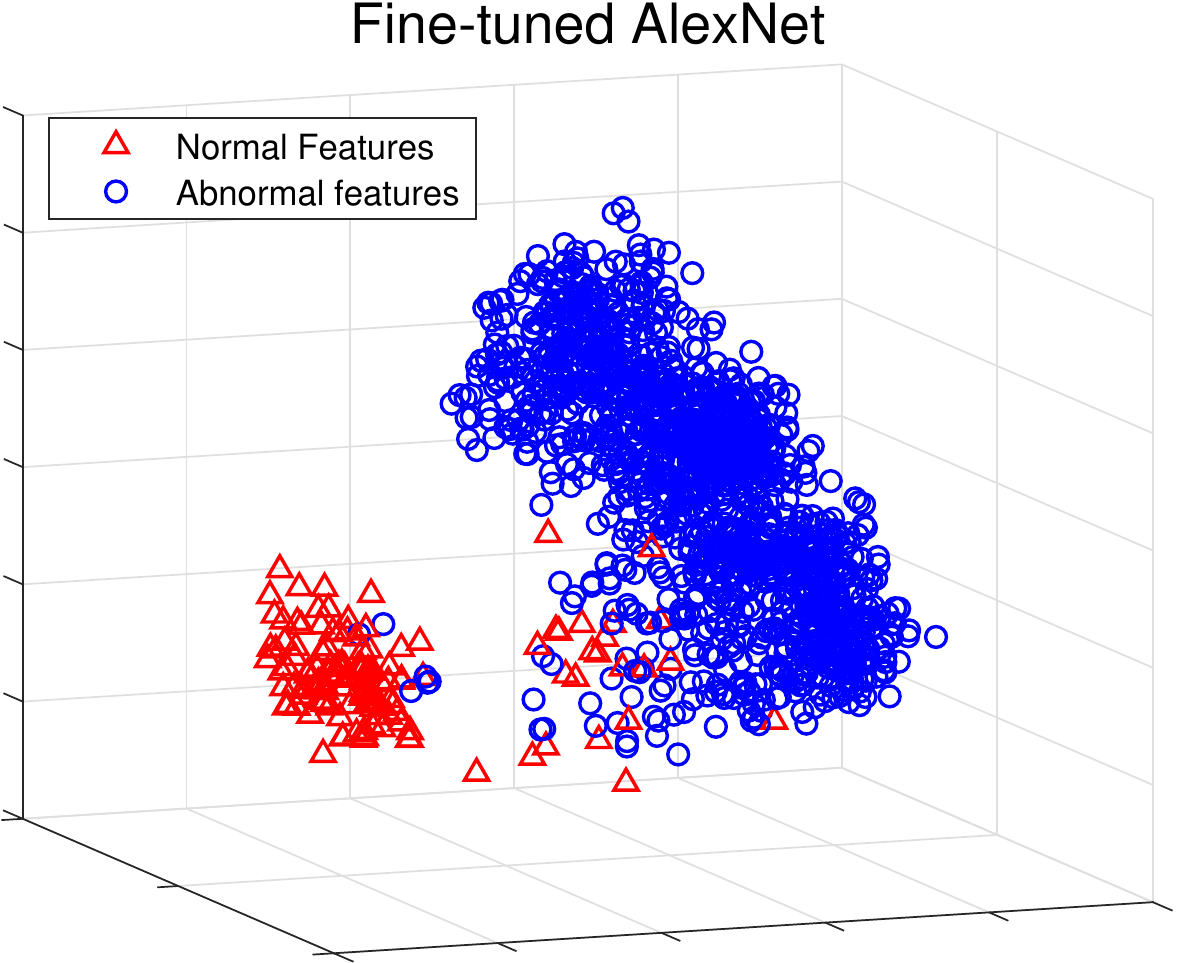}}
	\subfigure[]{ \centering
		\label{data091}
		\includegraphics[height=1.62in]{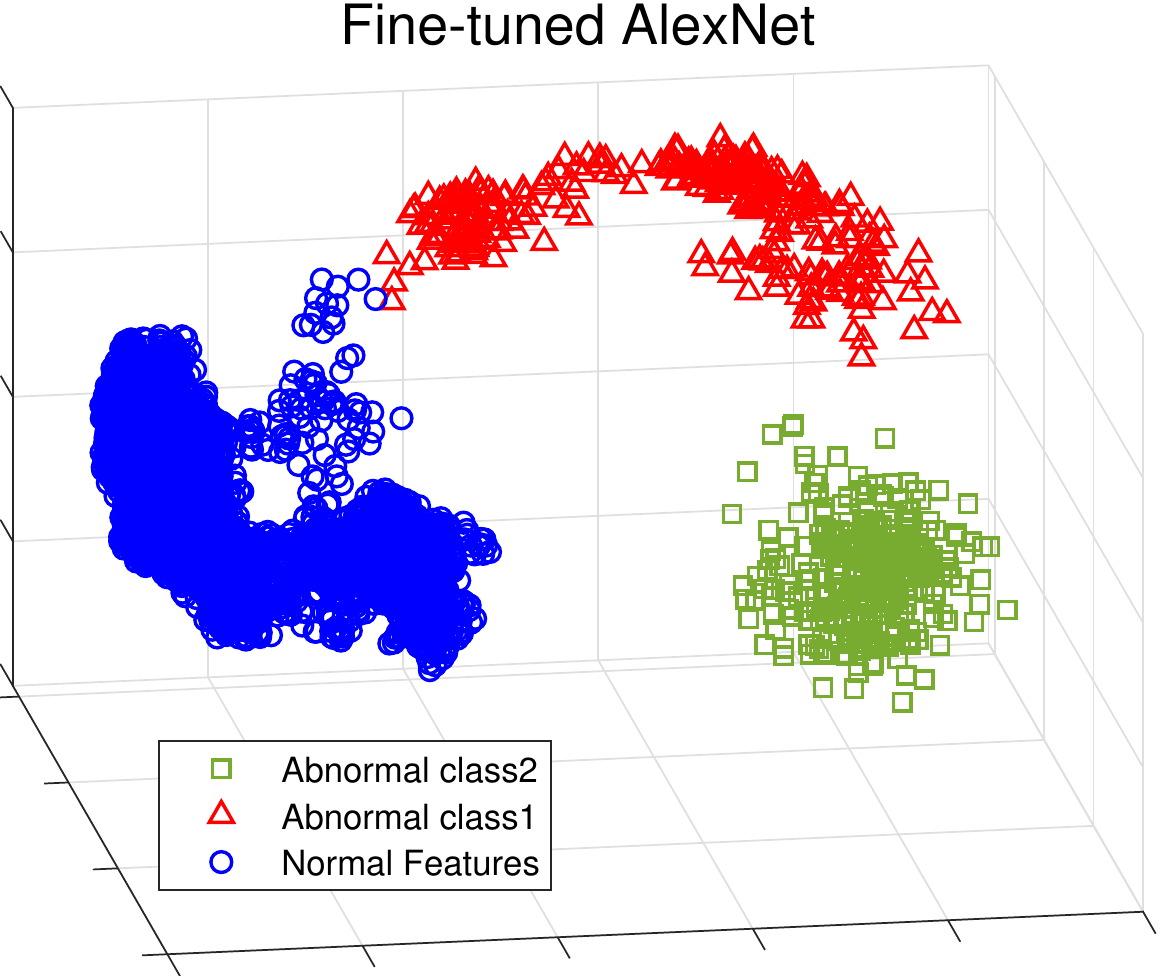}}
	\caption{This figure show some visualization of the normal and abnormal data in the feature space. The features extracted from the GPR images generated by the normal road is clearly separated from the features extracted from GPR images generated from other underground structures. Incremental OCSVM could gradually cluster the features in to classes, where the corresponding GPR image could be located and identified. Since the amount of normal data is much larger than that of anomalies, here we show the feature extraction and classification results of the data within a certain range around the anomalies.}
	\label{class_result}
\end{figure*}

The real-world experiments are conducted along 6 pavement roads, including 3 concrete roads and 3 asphalt roads. Along these roads, 15 GPR B-scan images are obtained by the GSSI-SIR30 GPR. The collected data has been manually classified, and some locations with part of underground anomalies were excavated and repaired, based on which the accuracy of the utilized methods could be measured. The horizontal length (i.e. detection distance) of each obtained image is greater than 30000 pixels, and the distance between two adjacent  pixel columns is 2cm.

Before applying the proposed method, three operations are conducted on the obtained images to eliminate the noise and highlight the subsurface objects: 1) First, the reflectance of the ground surface is eliminated in advance, which is supported by the Matgpr \cite{tzanis2006matgpr}; 2) Then, a filtering step based on the standard median filter is performed \cite{olhoeft2000maximizing} to reduce the electromagnetic noise and interferences; 3) Finally, concerning the compensation of the propagation losses caused by the medium attenuation and the signal energy radial dispersion, a nonlinear time-varying gain \cite{strange2002signal} is applied to the received signal. Since these three operations have been detailed and evaluated in our previous work\cite{zhou2018automatic}, they are not expanded in this paper.

\begin{figure}[h]
	\centering
	\subfigure[]{ \centering
		\label{normal_example}
		\includegraphics[width=0.465\textwidth, height=0.55in]{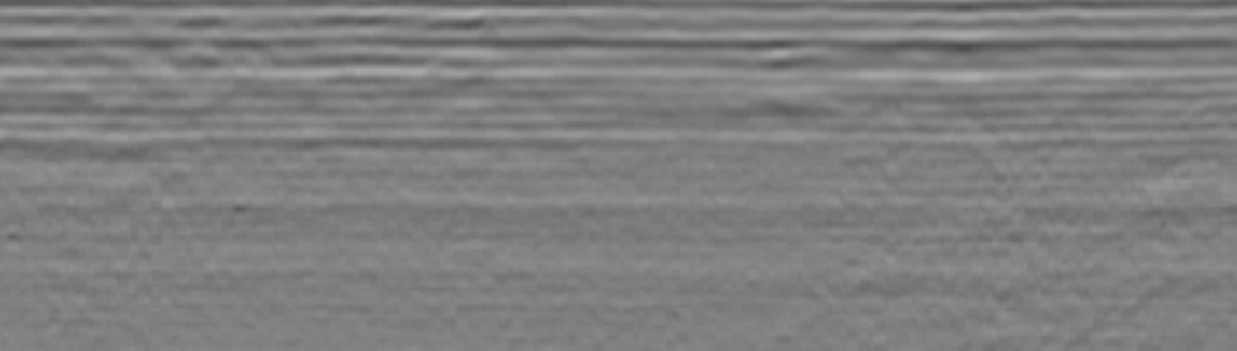}}
	\subfigure[]{ \centering
		\label{ab_example1}
		\includegraphics[width=0.465\textwidth, height=0.55in]{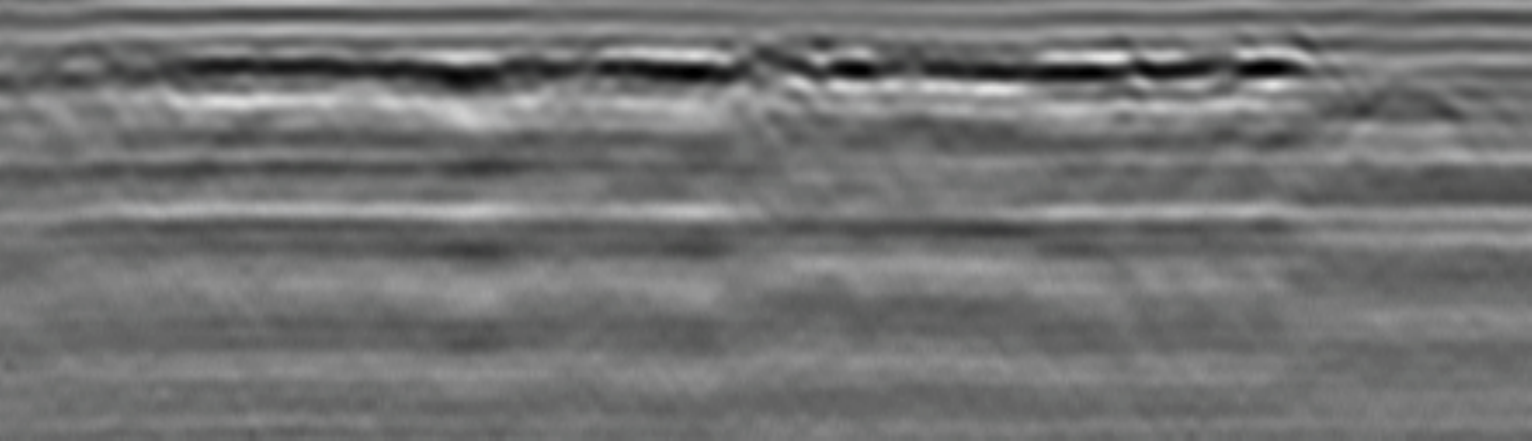}}
	\subfigure[]{ \centering
		\label{ab_example2}
		\includegraphics[width=0.225\textwidth, height=0.55in]{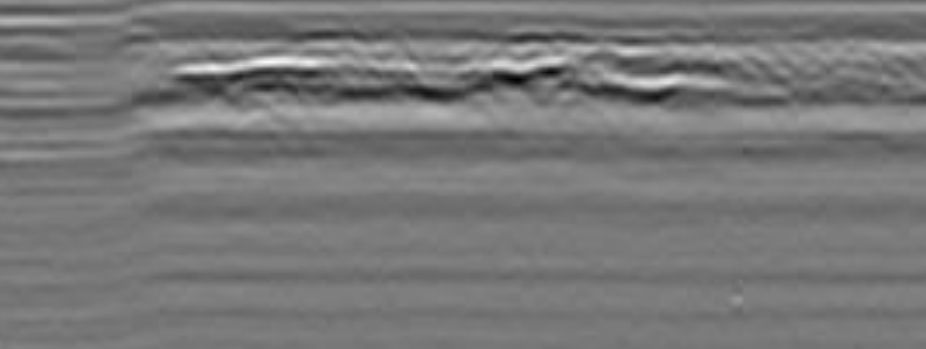}}
	\subfigure[]{ \centering
		\label{ab_example3}
		\includegraphics[width=0.225\textwidth,height=0.55in]{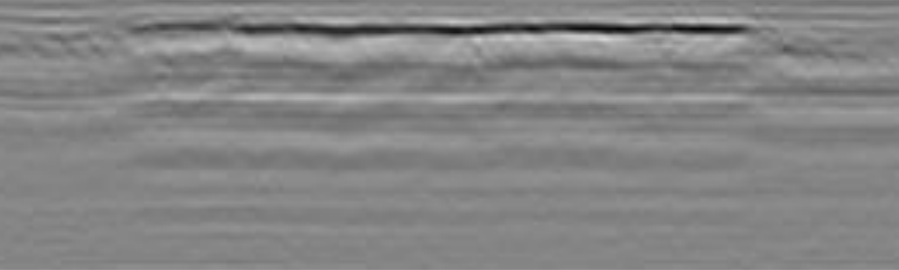}}		
	\subfigure[]{ \centering
		\label{ab_example4}
		\includegraphics[width=0.225\textwidth, height=0.55in]{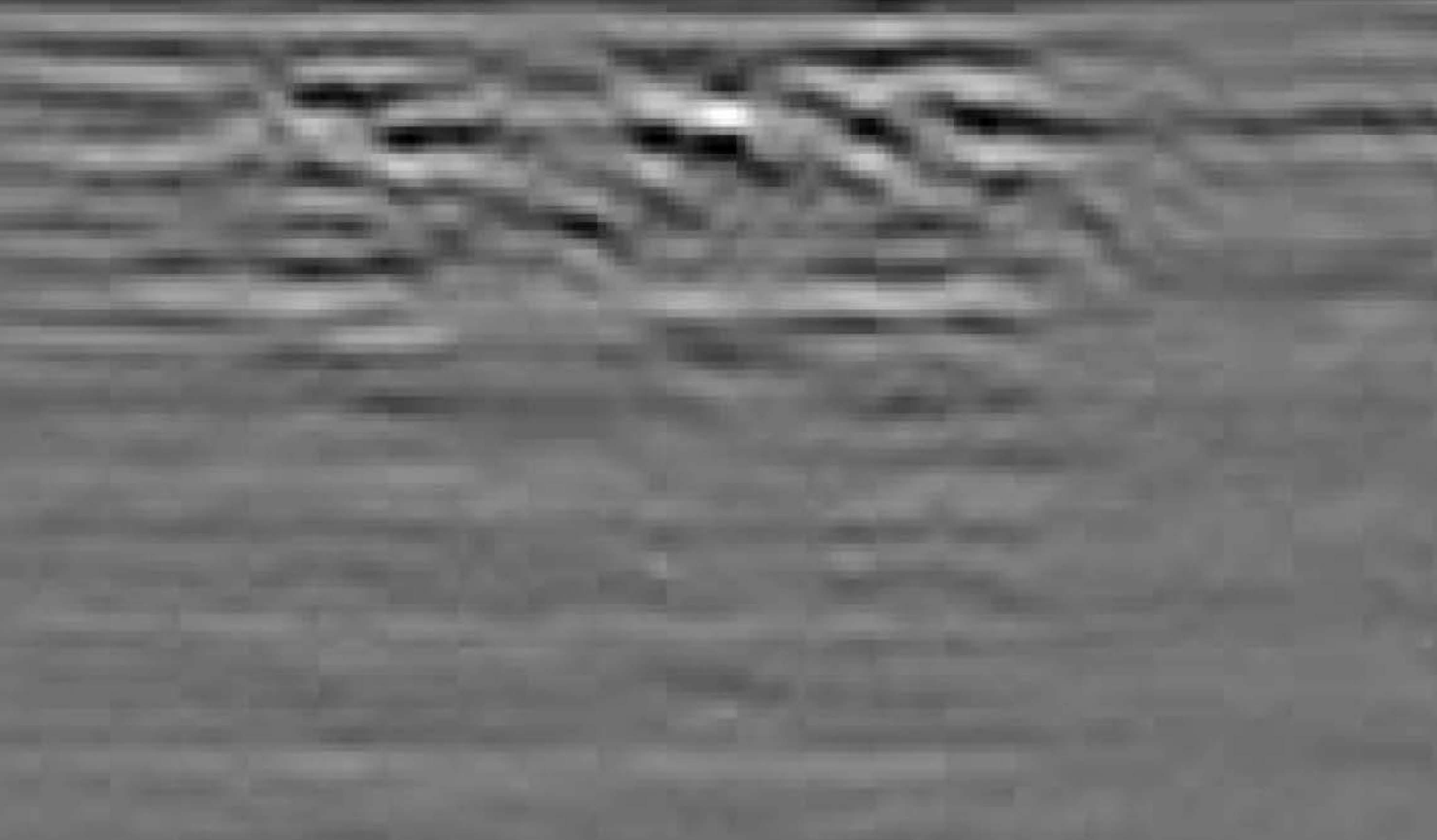}}	
	\subfigure[]{ \centering
		\label{ab_example5}
		\includegraphics[width=0.225\textwidth, height=0.55in]{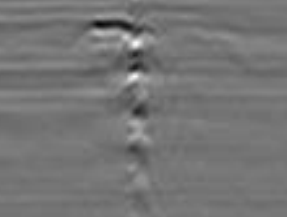}}
	\caption{(a) is the GPR B-scan image segment generated by a normal road without any underground structures. (b) to (e) illustrated some identified underground abnormalities  in the feature space. After on-site inspection, we determined that (b), (c), and (d) are generated by loose and hollow roadbeds, (e) is caused by rich of water, and (f) is a small crack (with the length less 50cm). These identified underground abnormalities  should be repaired to ensure the health of the detected roads.}
	\label{gprabnormal}
\end{figure}

For the obtained image, the section of the first 3000 to 4000 pixels in length without any subsurface objects is segmented to 300 normal GPR images. These images are then fused with the simulated images with underground objects to produce 300 synthetic images. After data augmentation, two types of images with more than 800 images each could be obtained.
As aforementioned,  considering the timeliness requirement,
the ResNet-18, VGG16, and AlexNet pre-trained on ImageNet are fine-tuned with the synthetic images. And the fine-tuned CNNs are then utilized to extract features of the rest of the image along this road with the sliding window of 300-pixel length and the swiped step of 10-pixel length.
Some results\footnote{It should be pointed out that the dimensions of the actual features are much larger than the three-dimensional. To show the classification intuitively, the Principal Component Analysis (PCA) \cite{abdi2010principal} is used to reduce the obtained features to three dimensions for visualization.} of the one-class learning the feature space are presented in Fig. \ref{class_result}. After completing the learning in the feature space, the corresponding overlapped GPR images in the sliding window that are classified in the same class are merged to obtain the final anomaly GPR images, since the length of some anomaly regions in the image exceeds the length of the sliding window.  Fig. \ref{gprabnormal} shows some of the merged images.

% Table generated by Excel2LaTeX from sheet 'Sheet1'
\begin{table}[htbp]
	\footnotesize
	\setlength{\tabcolsep}{4pt}
	\renewcommand{\arraystretch}{0.5}
	\centering
	\caption{Anomaly detection results with or without fine-tuning(\%)}
	\begin{tabular}{ccccccc}
		\toprule
		\multicolumn{7}{c}{Fine-Tuned} \\
		\midrule
		\multirow{2}[4]{*}{Road} & \multicolumn{2}{c}{ResNet-18} & \multicolumn{2}{c}{VGG16} & \multicolumn{2}{c}{AlexNet} \\
		\cmidrule{2-7}          & Precision & Recall & Precision & Recall & Precision & Recall \\
		\midrule
		1     & 94.24  & 95.32  & 94.37  & 94.72  & 90.29 & 98.12  \\
		\midrule
		2     & 94.12  & 96.26  & 92.42  & 95.43  & 92.59 & 95.71  \\
		\midrule
		3     & 93.43  & 99.01  & 95.13  & 96.71  & 91.41 & 94.31  \\
		\midrule
		4     & 94.21  & 97.21  & 94.11  & 95.31  & 93.17 & 95.12  \\
		\midrule
		5     & 95.27  & 95.19  & 94.02  & 96.89  & 92.39 & 95.21  \\
		\midrule
		6     & 93.29  & 95.14  & 92.19  & 96.18  & 90.11 & 94.67  \\
		\midrule
		Average & 94.09  & 96.36  & 93.71  & 95.87  & 91.66  & 95.52  \\
		\midrule
		&       &       &       &       &       &  \\
		\midrule
		\multicolumn{7}{c}{Not Fine-tuned} \\
		\midrule
		\multirow{2}[4]{*}{Road} & \multicolumn{2}{c}{ResNet-18} & \multicolumn{2}{c}{VGG16} & \multicolumn{2}{c}{AlexNet} \\
		\cmidrule{2-7}          & Precision & Recall & Precision & Recall & Precision & Recall \\
		\midrule
		1     & 82.26  & 84.36  & 90.41  & 82.53  & 81.29 & 87.03  \\
		\midrule
		2     & 87.21  & 82.14  & 82.46  & 86.73  & 79.21 & 84.11  \\
		\midrule
		3     & 86.38  & 83.26  & 83.45  & 83.22  & 83.47 & 79.48  \\
		\midrule
		4     & 89.82  & 89.21  & 89.21  & 90.04  & 88.02 & 78.23  \\
		\midrule
		5     & 88.27  & 84.29  & 84.23  & 84.27  & 84.11 & 85.17  \\
		\midrule
		6     & 85.29  & 82.14  & 81.19  & 82.28  & 87.21 & 82.67  \\
		\midrule
		Average & 86.54  & 84.23  & 85.16  & 84.85  & 83.89  & 82.78  \\
		\bottomrule
	\end{tabular}%
	\label{allresult1}%
\end{table}%

% Table generated by Excel2LaTeX from sheet 'Sheet4'
\begin{table}[htbp]
	\footnotesize
	\setlength{\tabcolsep}{2pt}
	\renewcommand{\arraystretch}{0.5}
	\centering
	\caption{Anomaly classification results with or without fine-tuning(\%)}
	\begin{tabular}{cccc}
		\toprule
		\multicolumn{4}{c}{Fine-tuned} \\
		\midrule
		Utilized CNN   & ResNet-18 & VGG16 & AlexNet \\
		\midrule
		Average Precision (Standard Deviation) & 85.1 (1.2)  & 82.3 (2.2) & 79.8 (3.1) \\
		\midrule
		Average Recall (Standard Deviation) & 82.0 (1.4)  & 84.1 (2.8)  & 81.9 (2.9)  \\
		\midrule
		&       &       &  \\
		\midrule
		\multicolumn{4}{c}{Not Fine-tuned} \\
		\midrule
		Utilized CNN   & ResNet-18 & VGG16 & AlexNet \\
		\midrule
		Average Precision (Standard Deviation) & 53.4 (4.1) & 49.2 (5.1)& 48.2 (5.9)\\
		\midrule
		Average Recall (Standard Deviation) & 39.3 (4.9) & 45.6 (4.8) & 45.1 (5.1)\\
		\bottomrule
	\end{tabular}%
	\label{allresult2}%
\end{table}%

For the task of anomaly detection, the effectiveness of the method could be measured from two levels: 1) whether the presence of anomalies can be detected, 2) and whether anomalies can be classified.
For detecting the presence of anomalies, the specific results of the proposed method are presented in Table \ref{allresult1}, along with the obtained results without fine-tuning.  For the classification of the subsurface anomalies, the results with or without fine-tuning are presented in Table \ref{allresult2}. 
It should be noted that in real-world applications, the actual types of the existing subsurface anomalies could be rarely determined without field excavation and restoration. It is also difficult to accurately define the classification of anomalies. For example, water-rich or voids could be classified into two categories, but if there is stagnant water in the voids, there is controversy. Since the anomalies have been surveyed in the experimental areas of this paper as aforementioned, the existing anomalies could be classified\footnote{For abnormal classification in other areas, the classification results and basis of this paper might not be portable.}: underground cavity, loose, fault (more than one meter in length), rich in water (including water in the cavity). From the practical engineering significance, the experimental results of the first level of detecting the presence of anomalies could be more meaningful. From the results, whether it is anomaly detection or classification, the method proposed in this paper has achieved considerable results. Specific analysis would be presented in the following

The used fine-tuning times of the utilized CNNs are presented in Table \ref{ftime}. The utilized networks converge within 20 iterations, and the training time is within 5 minutes, ensuring the timeliness of the method in practical applications. 

\begin{table}[htbp]
	\footnotesize
	\renewcommand{\arraystretch}{0.8}
	\centering
	\caption{Average Fine-tuning time of utilized CNNs}
	\begin{tabular}{cccc}
		\toprule
		Utilized  CNN & ResNet-18 & VGG16 & AlexNet \\
		\midrule
		Fine-tuning Time & 57s   & 149s  & 45s \\
		\bottomrule
	\end{tabular}%
	\label{ftime}%
\end{table}%

\subsection{Analysis of Conducted Experiments}

From the experimental results (Table \ref{allresult1}), the utilized CNNs only pre-trained on ImageNet achieve over 80\% detection precision and recall on GPR images, where the effectiveness of CNNs as feature extractors could be demonstrated to some extend. Meanwhile, fine-tuning pre-trained CNNs with synthetic data has improved both the precision and recall for identifying anomalies in GPR images. In particular, the recall rate increased by about 10 percentage points. This is of practical value, since missing anomalies would have a negative impact on urban security. If some anomalies are not identified in early, the follow-up could cause more serious effects, such as road collapse, underground cavity, etc.

\begin{figure}[htpb]
	\centering
	\subfigure[]{ \centering
		\label{not_identified1}
		\includegraphics[width=0.48\textwidth,height=0.68in]{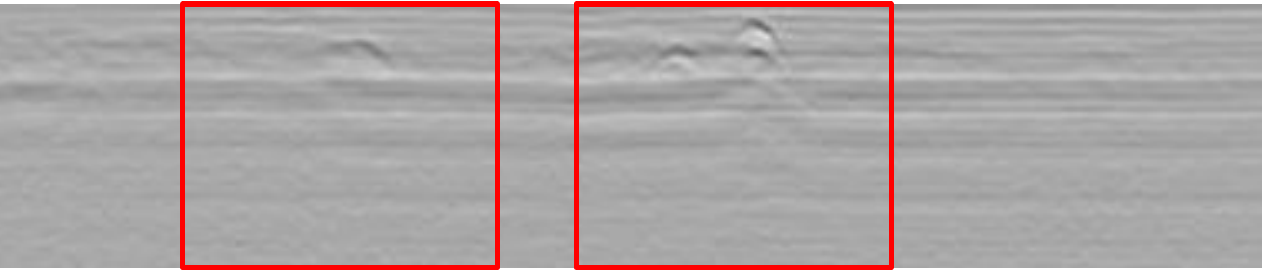}}
	\subfigure[]{ \centering
		\label{not_identified2}
		\includegraphics[width=0.48\textwidth,height=0.68in]{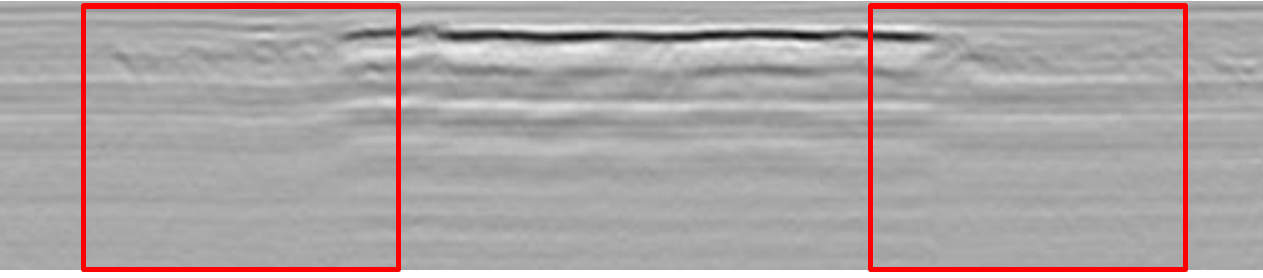}}
	\caption{(a) and (b) show two examples that could be misidentified. (a) shows a small subsurface object whose dielectric constant does not differ much from the surrounding environment. To improve the data processing efficiency, this paper processes the GPR images obtained from a certain area in batches, so the characteristics of such objects on the GPR images are not highlighted. (b) shows the moment when the sliding window has just slipped or slid across a subterranean anomaly. At this time, the proportion of anomalies in the sliding window is small and may be ignored.}
	\label{not_identified}
\end{figure}

Based on the identified features, the corresponding GPR image could be segmented, and the subsurface anomalies could be located. By checking the results with or without fine-tuning, it could be found that the subsurface anomalies that are not identified by the CNN without fine-tuning are mainly small anomalies, or when the sliding window is just starting to slide from normal to abnormal. Fig. \ref{not_identified} illustrates two examples of these kinds of GPR images. To improve the data processing efficiency, the GPR images obtained from a certain area are processed in batches through the above three operations, so the characteristics of small objects on the GPR images are not highlighted manually, causing challenges for CNNs to recognize such objects. For the case where the sliding window has just entered or drawn out the abnormal area, we confirm that this position should also be marked as an anomaly, since there will be errors using GPS or odometer when performing abnormal positioning, and the appropriate redundancy could improve robustness to avoid omissions.
After fine-tuning, the above two kinds of omissions are significantly reduced and recall improved.
\begin{figure}[htbp]
	\centering
	\subfigure[]{ \centering
		\label{gprtest1}
		\includegraphics[width=0.23\textwidth]{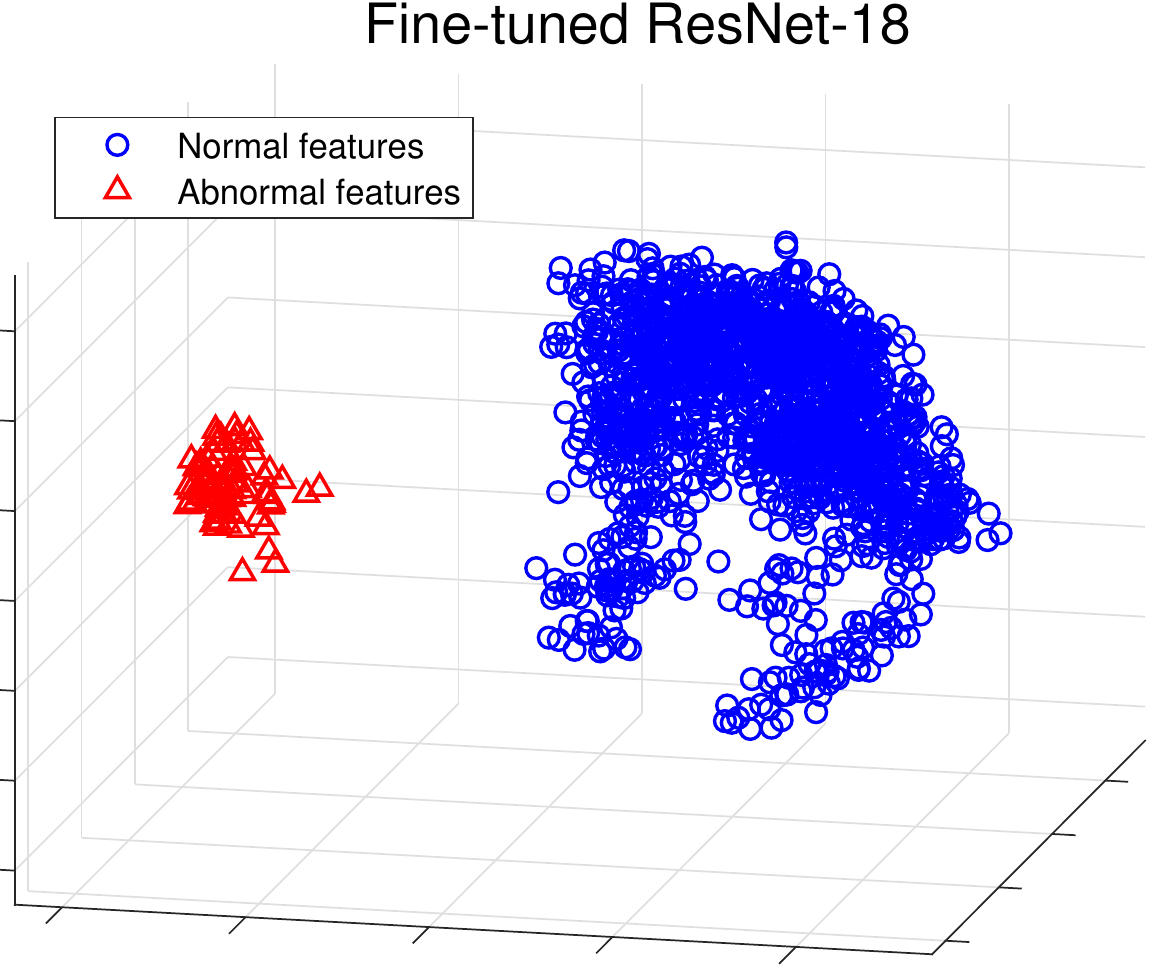}}
	\subfigure[]{ \centering
		\label{gprtest2}
		\includegraphics[width=0.23\textwidth]{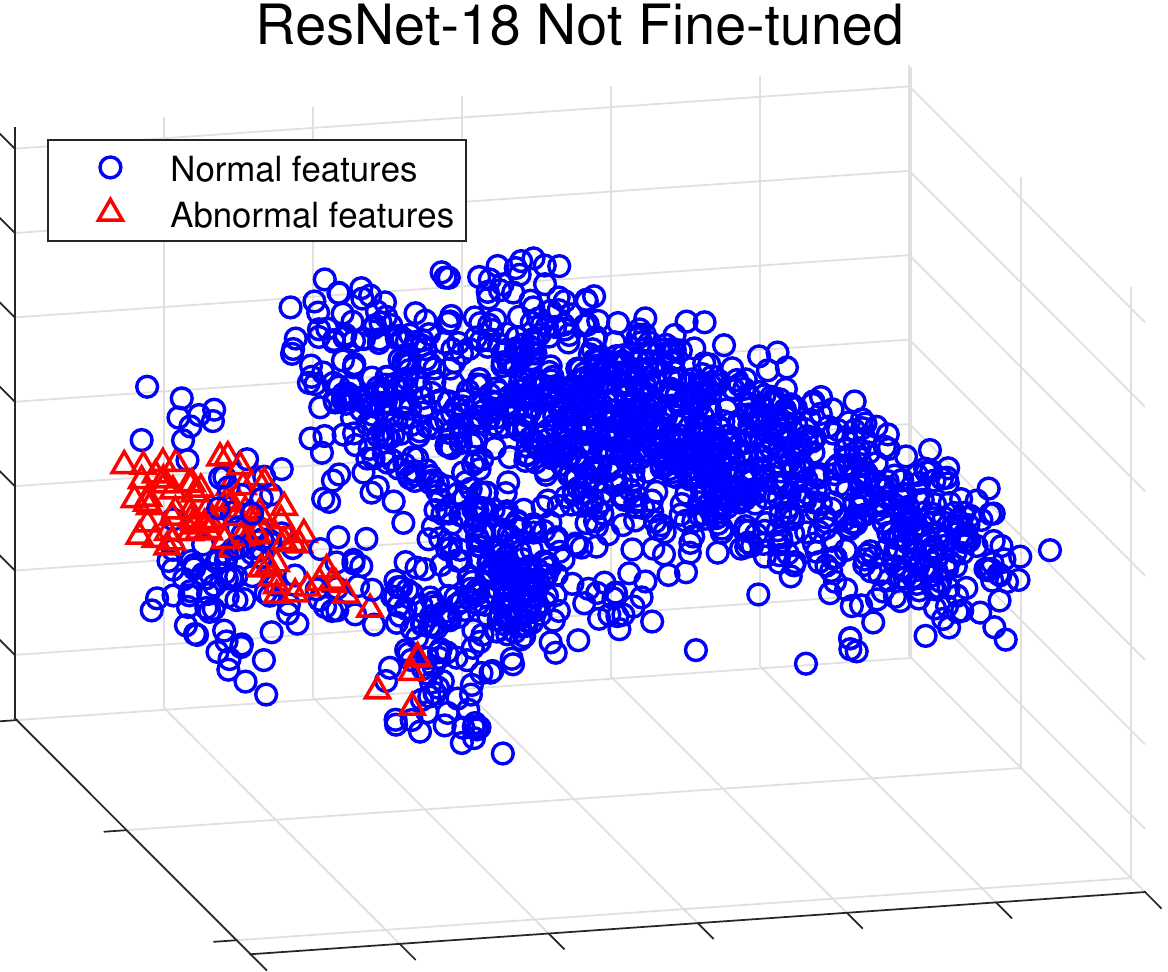}}
	\caption{The extracted features of an anomaly with and without fine-tuning. It can be intuitively observed that the fine-tuning makes the inter-class distance significantly larger}
	\label{fcompare}
\end{figure} 
Fig. \ref{fcompare} shows the extracted features of a small anomaly and the surrounding normal images by ResNet-18 with or without fine-tuning. It could be observed that the fine-tuned network can better distinguish the small anomaly, i.e. the red and blue data points have greater inter-class distances. Actually, in the process of feature extraction by recent CNNs, without any data or training constraints, some detailed characteristics on the image could be discarded in the process such as pooling or spaced convolution. During fine-tuning, we used anomalous images of different sizes for training (including small anomalies), forcing the network to enhance its ability to recognize such features.

From the above experiments, it could be observed that the proposed fine-tuning could effectively improve the feature extraction of CNNs pre-trained on visual images. 
To validate the value of the synthetic data, we also fine-tuned and compared the utilized networks with several different data sources:
\begin{enumerate}
	\item The real data obtained and labeled in the conducted experiments. Since the experiment contains 3 cement roads and 3 asphalt roads, we select each one of them for manual labeling to obtain the labeled data.
	\item The proposed synthetic data and normal data.
	\item Simulated and normal data. The simulated data is generated by GprMax without fusion.
	\item Using pre-trained CNNs for feature extraction without fine-tuning.
\end{enumerate}

\begin{figure}[htbp]
	\centering
	\subfigure[]{ \centering
		\label{pre}
		\includegraphics[width=0.45\textwidth]{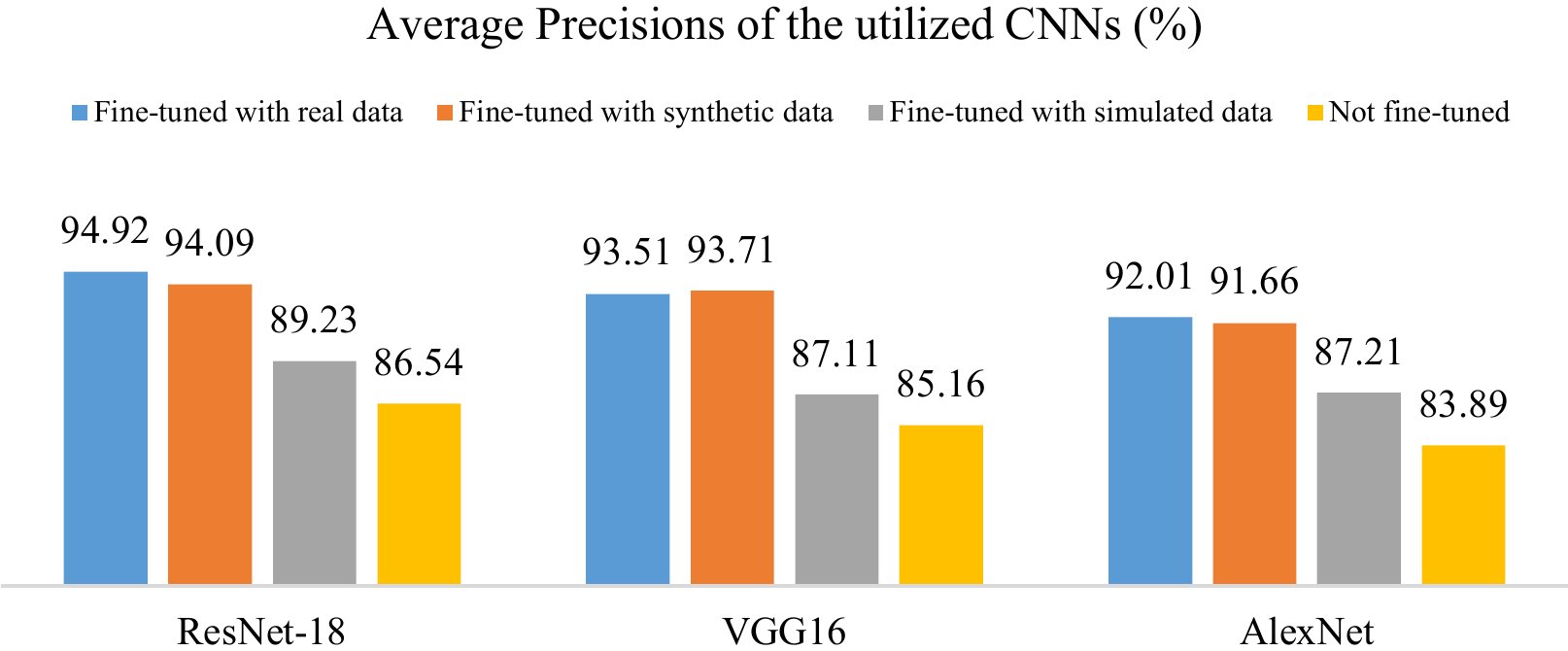}}
	\subfigure[]{ \centering
		\label{rec}
		\includegraphics[width=0.45\textwidth]{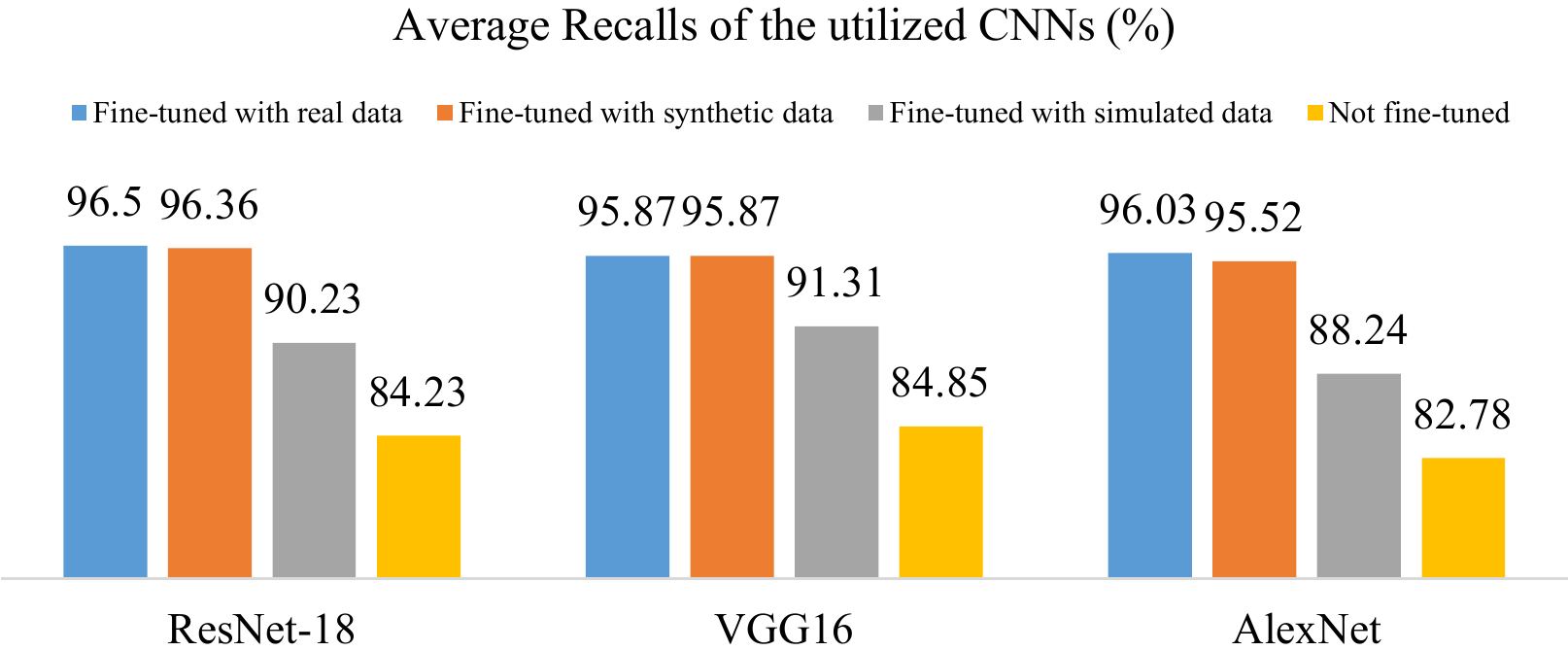}}
	\caption{(a) and (b) show the average detection precisions and recalls of the utilized ResNet-18, VGG, and AlexNet in the conducted experiments under different fine-tuning settings.}
	\label{recall}
\end{figure} 

The average anomaly detection precisions and recalls of different settings are shown in Fig. \ref{recall}. 
It could be observed that using normal and simulated data can improve both the precision and recall, but the effect is not comparable to fine-tining with synthetic data. Fine-tuning with real data does achieve good precision and recall, but the time spent on labeling the data cannot be ignored. The use of synthetic data can not only achieve better results, but also effectively reduce the time and workload of manual data annotation.
In fact, the synthesized image contains not only the general subsurface environment characteristics of the certain area, but also the characteristics of various subsurface objects. Similar to visual images, synthetic images actually use the general environment as the background. Fine-tuning with synthetic and normal data could improve the network's ability to extract features in this specific background. Moreover, during real-world detection, the method in this paper could be used to firstly analyze the acquired data. And with the increase of acquired data, the classified real data can be segmented and gradually added to the fine-tuning process to continuously improve the feature extraction of the utilized CNNs. Besides, Pre-training and GprMax data generation do not need to be performed on-site, which also improves the efficiency of on-site detection.

\section{Conclusion}

The subsurface conditions of an area could be imaged through the use of GPR, while automatically identifying the underground objects of the detection area in the obtained GPR image could be challenging.
In this paper, a novel method based on CNN and one-class learning is proposed to locate and identify anomaly objects in GPR images. After detecting an area using GPR, a collected normal (i.e. without subsurface objects) GPR image section is segmented and fused with simulated GPR image to generate some synthetic data. A pre-trained CNN is then fine-tuned by the generated synthetic data to capture GPR image features of normal and abnormal underground objects in the detection area. 
After that, the obtained GPR image is swiped with a sliding window, and the fine-tuned CNN is utilized to extract features of the GPR image in each window. These extracted features could finally be classified by the one-class learning algorithm in the feature space, thus the corresponding underground objects are identified and located. 

Experiments on real-world datasets are conducted. From the experimental results, it could be observed that fine-tuning the pre-trained CNN with synthetic data specifically designed for the detection area could significantly improve the feature extraction effect in identifying underground objects. We could also found that the quality of the training data has a significant impact on the performance of CNNs. The fact that one-class learning does not require preset categories is also applicable to the detection of unknown underground environments. Meanwhile, the fine-tuning time of the utilized CNN is also considerable, ensuring the timeliness of the proposed method in real-world applications. 

There are still some points need to pay attention to. 1) The GPR image is a kind of data that relates both horizontally and vertically, which not only has the continuity of detection time or position in the horizontal direction but also correlates in the vertical direction due to the continuity of the underground
medium. In future work, we hope to modify networks around this point to improve the efficiency of feature extraction. 2) On the other hand, differences exist in the GPR images obtained in different areas and environments, or by different types of utilized devices. In future work, we intend to enrich the GPR image database by crowd sourcing annotation.

\bibliography{ref}
\bibliographystyle{IEEEtran}

\end{document}